\documentclass[journal]{IEEEtran}
\usepackage{amsmath,amsfonts}
\usepackage{algorithmic}
\usepackage{algorithm}
\usepackage{array}
\usepackage[caption=false,font=normalsize,labelfont=sf,textfont=sf]{subfig}
\usepackage{textcomp}
\usepackage{stfloats}
\usepackage{url}
\usepackage{verbatim}
\usepackage{graphicx}
\usepackage{cite}
\usepackage{graphicx}
\usepackage{multirow}
\usepackage{overpic}
\usepackage{amsmath}
\usepackage{xcolor}
\usepackage{colortbl}
\usepackage{amssymb}
\usepackage{pgfplots}
\usepackage{booktabs}
\pgfplotsset{compat=1.18}
\definecolor{myblue}{RGB}{31,119,180}
\definecolor{mygreen}{RGB}{44,160,44}
\definecolor{myred}{RGB}{214,39,40}

\begin{document}

\title{DACL-RAG: Data Augmentation Strategy with Curriculum Learning for Retrieval-Augmented Generation}

\author{Shaohan Wang,~Licheng Zhang,~Zheren Fu,~Zhendong Mao,~\IEEEmembership{Member,~IEEE}, and~Yongdong Zhang,~\IEEEmembership{Fellow,~IEEE}
\thanks{Shaohan Wang, Licheng Zhang and Zheren Fu are with the School of Information Science and Technology, University of Science and Technology of China, Hefei, Anhui 230022, China (e-mail: wsh2000@mail.ustc.edu.cn; zlczlc@mail.ustc.edu.cn; fzr@mail.ustc.edu.cn).}
\thanks{Zhendong Mao and Yongdong Zhang are with the School of Information Science and Technology, University of Science and Technology of China, and the Institute of Artificial Intelligence, Hefei Comprehensive National Science Center, Hefei, Anhui 230022, China (e-mail: zhyd73@ustc.edu.cn; zdmao@ustc.edu.cn).}

}

\markboth{Journal of \LaTeX\ Class Files,~Vol.~14, No.~8, August~2021}%
{Shell \MakeLowercase{\textit{et al.}}: A Sample Article Using IEEEtran.cls for IEEE Journals}


\maketitle

\begin{abstract}
Retrieval-Augmented Generation (RAG) is an effective method to enhance the capabilities of large language models (LLMs). 
Existing methods typically optimize the retriever or the generator in a RAG system by directly using the top-$k$ retrieved documents. 
However, two key issues inherent in the training data constrain the effectiveness of this training paradigm: (1) across different queries, the top-$k$ retrieved documents vary greatly in content quality, with some providing valuable knowledge while others lack critical information or are even misleading, and training on such data in a purely random manner may impair the generator’s ability to extract key information; (2) for a given query, the limited set of $k$ documents often exhibits low discriminability, and training solely on them makes it difficult for the retriever to learn how to distinguish between relevant and irrelevant documents.
To address these issues, we introduce \textbf{DACL-RAG}, a multi-stage RAG training framework that combines a multi-level \textbf{D}ata \textbf{A}ugmentation strategy with a multi-stage \textbf{C}urriculum \textbf{L}earning paradigm. 
The data augmentation strategy constructs comprehensive and diverse training sets with controllable difficulty levels  through sample evolution, 
while the curriculum learning paradigm organizes them into progressive stages for training, ensuring stable and consistent improvements, thereby optimizing the overall performance and generalization of the RAG system more effectively. 
Our DACL-RAG framework demonstrates consistent effectiveness across four open-domain QA datasets, achieving performance gains of 2\% to 4\% over multiple advanced methods.
\end{abstract}


\section{Introduction}
\begin{figure}[t]
  \includegraphics[width=1\linewidth]{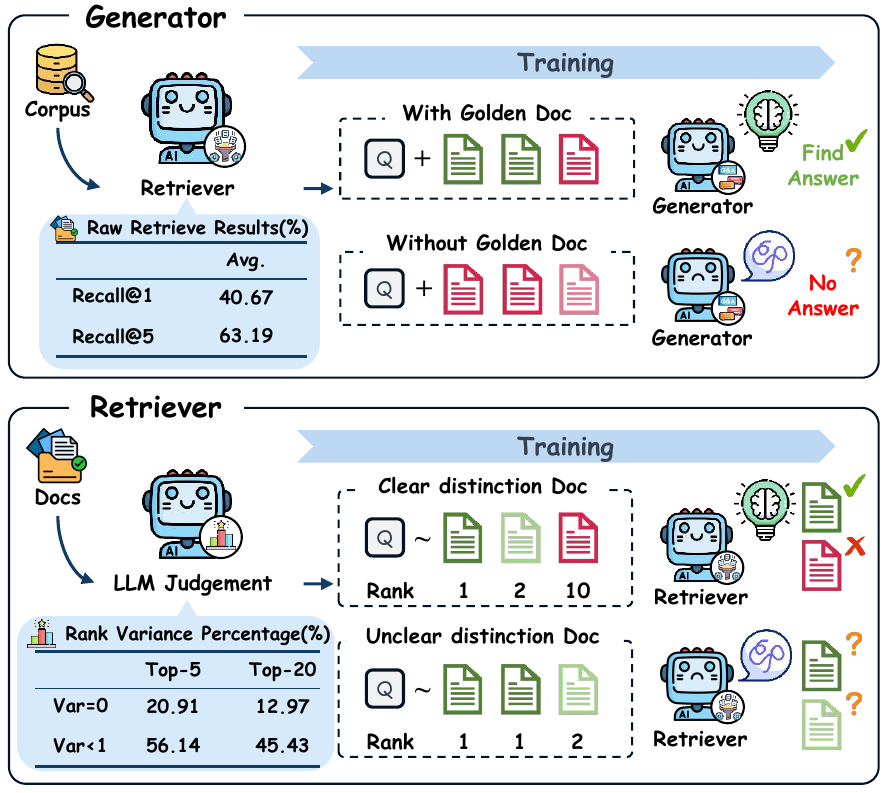}
  \caption {\textbf{Existing training paradigm.} Here, \textcolor{mygreen}{green} denotes documents that support the model’s responses, while \textcolor{red}{red} denotes documents that are useless or even harmful. We conduct analysis on four open-domain QA datasets: NQ, TriviaQA, HotpotQA, and PopQA. For generator training (upper panel): \textit{Raw Retrieve Result} reports the answer recall of the original retrieved documents; these raw retrievals exhibit low recall of answer-bearing evidence, with some documents containing the answer and others not, so training directly on such inputs injects noise and impedes optimization. For retriever training (lower panel): after scoring document usefulness with an LLM, we find that within the $top$-$5$ candidates the ranking scores exhibit limited dispersion, providing insufficient signal to learn robust discrimination between helpful and unhelpful documents; expanding the sampling window to the $top$-$20$ increases score separation and mitigates this issue.}
  \label{fig:moti}
\end{figure}

\IEEEPARstart{L}{arge} language models (LLMs) have demonstrated remarkable capabilities in a wide range of Natural Language Processing (NLP) tasks~\cite{brown2020language, anil2023palm, dubey2024llama}, but they are still constrained by the limitations of the static knowledge embedded within their internal parameters~\cite{roberts2020much, kandpal2023large, Gao2023RetrievalAugmentedGF}.
Retrieval-Augmented Generation (RAG) addresses this limitation by supplementing LLMs with additional knowledge retrieved from external knowledge bases, and has significantly enhanced the capabilities of existing large models in tasks such as Open-Domain Question Answering~\cite{10.5555/3495724.3496517, min-etal-2020-ambigqa, lewis-etal-2020-bart, izacard2023atlas, shi2023replug, yoranmaking, linra, Fang2024EnhancingNR, ke-etal-2024-bridging, xu2024recomp, huang2024raven} and Dialog System~\cite{wang2024unims, liu2024chatqa, wang2024retrieval}.

The overall performance of the RAG system depends crucially on the quality of the retrieved documents and the LLMs' ability to effectively utilize them.
Accordingly, much effort has been devoted to enhancing RAG systems from two main directions: (1) improving the retriever by leveraging LLMs to assess document utility and provide supervision signals~\cite{shi2023replug, zhang2024multi, zhang-etal-2024-arl2, ke-etal-2024-bridging, xu2024recomp}; (2) training Retrieval-Augmented Language Models (RALMs) that can better utilize retrieved documents~\cite{10.5555/3495724.3496517, min-etal-2020-ambigqa, lewis-etal-2020-bart, izacard2020leveraging, Fang2024EnhancingNR, yoranmaking, NEURIPS2024_db93ccb6}, or combining both~\cite{li2025ragddroptimizingretrievalaugmentedgeneration, linra} to enhance the overall performance. These methods generally employ an off-the-shelf retriever to obtain the $top$-$k$ most relevant documents for each query, which are then combined with the corresponding query to train either the retriever or the generator separately.
However, the effectiveness of this training paradigm is constrained by two key issues stemming from the inherent characteristics of the training data.
(1) Across different queries, the $top$-$k$ retrieved documents vary greatly in content quality, which introduces considerable noise into generator training, as shown in Fig.~\ref{fig:moti} (top panel). Among the retrieved documents, the $top$-$5$ answer recall is 63.19\%, meaning that many queries still have no answer in the top five retrieved documents, reflecting substantial noise and distractors. When such documents are randomly used during training, they degrade the learning process of the generator, hindering its ability to effectively extract and utilize relevant information.
(2) For a given query, the $top$-$k$ documents often have comparable effectiveness in supporting the answer, thereby limiting informational diversity and weakening the supervision signal for retriever training. 
As shown in Fig.~1 (bottom panel), for over 50\% of queries, the variance of answer-utility scores across the top-5 documents is less than 1, indicating limited dispersion and thereby making it difficult for the retriever to learn fine-grained distinctions in document contribution.
A few recent studies have attempted to address these issues by synthesizing different types of hard-to-learn noise to improve the robustness of the generator~\cite{yoranmaking, Fang2024EnhancingNR}. However, this approach neglect the importance of clean data, which is essential for enabling RALMs to extract and utilize relevant information effectively, and offer no benefit toward retriever optimization.
Thus, how to construct a balanced and systematic training strategy to improve the learning of both retriever and generator remains underexplored and deserves more attention.

Curriculum learning~\cite{bengio2009curriculum} provides an effective data-organization scheme in which models begin with easy examples and gradually tackle more difficult ones, thereby building the capacity to cope with noise and ambiguous samples.
Building on this principle, we introduce \textbf{DACL-RAG}, a novel RAG training framework that combines a multi-level \textbf{D}ata \textbf{A}ugmentation strategy with a multi-stage \textbf{C}urriculum \textbf{L}earning paradigm.
The data augmentation strategy constructs comprehensive and diverse training sets with controllable difficulty levels, 
while the curriculum learning paradigm organizes them into progressive stages for training, ensuring stable and consistent improvements.
Specifically, for the generator, difficulty is defined by the noise ratio within documents. By rewriting golden documents to enhance query information or craft counterfactual distractor documents, we construct multiple training datasets with gradually increasing noise ratios. 
For retriever training, difficulty is defined based on the discriminability in utility between different documents for answering a given query. 
The smaller the discriminability in utility, the higher the difficulty, as the limited separability provides insufficient signal for the retriever to learn to distinguish useful from unhelpful documents.
Compared to existing methods that typically select only the top-ranked documents, we first broaden the overall sampling range to include a wider variety of documents. A well-trained RALM is used to evaluate utility of documents and re-rank them, and we construct training sets from easy to difficult by gradually reducing the ranking gap between sampled documents.

As a result, DACL-RAG bring two advantages to RAG system: (1) For the generator, DACL-RAG systematically transitions it from merely extracting information to effectively countering potential distracting noise within documents. (2) For the retriever, DACL-RAG train it in stages to progressively distinguish documents with obvious differences and then distinguish those with only slight differences. 
This progressive training framework has led to an overall performance and generalization improvement of the RAG system.
Our contributions can be concluded as follows:
\begin{itemize}
\item We propose the DACL-RAG training framework for RAG system based on the concept of curriculum learning. To the best of our knowledge, this is one of the first times that CL strategy has been applied to optimize RAG systems.
\item We defined the task-specific difficulty levels for the retriever and generator in RAG, and designed a complete method for constructing curriculum-structured training sets.
\item We evaluated our DACL-RAG framework on four popular datasets, demonstrating consistent performance gains of 2\% to 4\% over multiple advanced methods, thereby highlighting the superiority of our approach.
\end{itemize}

\section{Related Work}
\subsection{Retrieval-augmented Generation}
Using documents retrieved from extended knowledge bases to enhance the capabilities of large language models (LLMs) has been proven effective in NLP tasks, including language modeling~\cite{ram2023context, zhang2024multi} and question answering~\cite{10.5555/3495724.3496517, min-etal-2020-ambigqa, lewis-etal-2020-bart, izacard2023atlas, shi2023replug, yoranmaking, linra, Fang2024EnhancingNR, ke-etal-2024-bridging, huang2024raven}.
Specifically, a Retrieval-Augmented Generator (RAG) system takes a query as input and uses a retriever to retrieve relevant documents from an external knowledge base. Then, it combines the documents with the query and feeds them into the LLM to make up for it's own lack of knowledge.

Optimization of the RAG system focuses on two main areas: improving the retriever and enhancing the generator (LLM) to the RALM. On the retriever side, Replug~\cite{shi2023replug} uses KL divergence to align retriever results with generator's preferences. ARL2~\cite{zhang-etal-2024-arl2} uses LLM to directly label evidence support and combines list-wise and pairwise ranking losses with adaptive self-training. LLM-Embedder~\cite{zhang2024multi} employs a distillation objective based on model rankings. SuRe~\cite{kim2024sure} performs evidence summarization and assessment over multiple candidate answers to choose the model’s most endorsed output. RECOMP~\cite{xu2024recomp} trains a compressor to summarize documents returned by the retriever, thereby providing more suitable context for the generator. BGM~\cite{ke-etal-2024-bridging} trains an intermediate model that aligns retrieved evidence with the generator’s preferences. In general, these approaches either optimize the retriever directly or employ an intermediate model to better match the generator’s preferences.

For the generator, RAG~\cite{10.5555/3495724.3496517}, AmbigQA~\cite{min-etal-2020-ambigqa} and BART~\cite{lewis-etal-2020-bart} introduced retrieval augmented generative models for open domain question answering. FiD~\cite{izacard2020leveraging} finetunes the generator to handle retrieved documents and queries, addressing irrelevant information. LUMEN~\cite{10.5555/3618408.3618698} adds an offline–online memory encoding scheme for FiD, whereas Glimmer~\cite{dejong2023glimmergeneralizedlateinteractionmemory} further adds a shallow reranker over the pre-computed memory. Self-RAG~\cite{asai2024selfrag} trained the generator to choose whether and when to retrieve, leading to better answers. RankRAG~\cite{NEURIPS2024_db93ccb6} instruction-tunes a single generator to perform both context reranking and answer generation, improving retrieval–generation consistency. Other studies introduce noise to improve model robustness~\cite{yoranmaking, Fang2024EnhancingNR}.

Combining the strengths of both approaches, RA-DIT~\cite{linra} uses modular training to optimize the retriever and generator separately, enhancing overall RAG system performance. RAG-DDR~\cite{li2025ragddroptimizingretrievalaugmentedgeneration} uses an end-to-end loop of sampling, scoring, and training to alternately update the refinement and generation modules, align their preferences, and improve RAG performance.
\subsection{Curriculum Learning}
Curriculum Learning (CL)~\cite{bengio2009curriculum} is a machine learning strategy that mimics human learning by training models gradually, starting with simpler tasks and progressing to more complex ones. It aims to improve the model's generalization ability and accelerate its convergence speed~\cite{10.1007/s11263-022-01611-x}. Early studies have extensively investigated CL in computer vision domain and have demonstrated its advantages in training deep models~\cite{guo2018curriculumnet, hacohen2019power, chen2023multi, ranjan2017curriculum}. In the NLP domain, \cite{platanios-etal-2019-competence} adopt a heuristic approach to difficulty ordering, building the curriculum using sentence length and the rarity of vocabulary, however, empirical evidence shows that such heuristics yield only limited gains~\cite{surkov-etal-2022-data}. Beyond such heuristic annotation schemes, \cite{xu2020curriculum} systematically explore and validate Curriculum Learning (CL) for fine-tuning language models on Natural Language Understanding (NLU) tasks by leveraging external teacher models, achieving consistent gains across Machine Reading Comprehension (MRC), Natural Language Inference (NLI), and related benchmarks. Building on this line, \cite{feng-etal-2025-pretrained} leverage the PLM itself to estimate sample difficulty and prioritize fine-tuning examples accordingly, improving both convergence speed and final performance. Recent work has further applied the CL paradigm to specialized domains such as code~\cite{nair-etal-2024-curriculum} and knowledge graphs~\cite{wang-etal-2025-knowledge}. For retrievers, recent work demonstrates that progressively increasing the difficulty of sampled data likewise yields substantial improvements when training embedding-based models~\cite{Zhu2022FromET,zeng2022curriculum,he2023capstone}.

\begin{figure*}[t]
  \centering
  \includegraphics[width=1\linewidth]{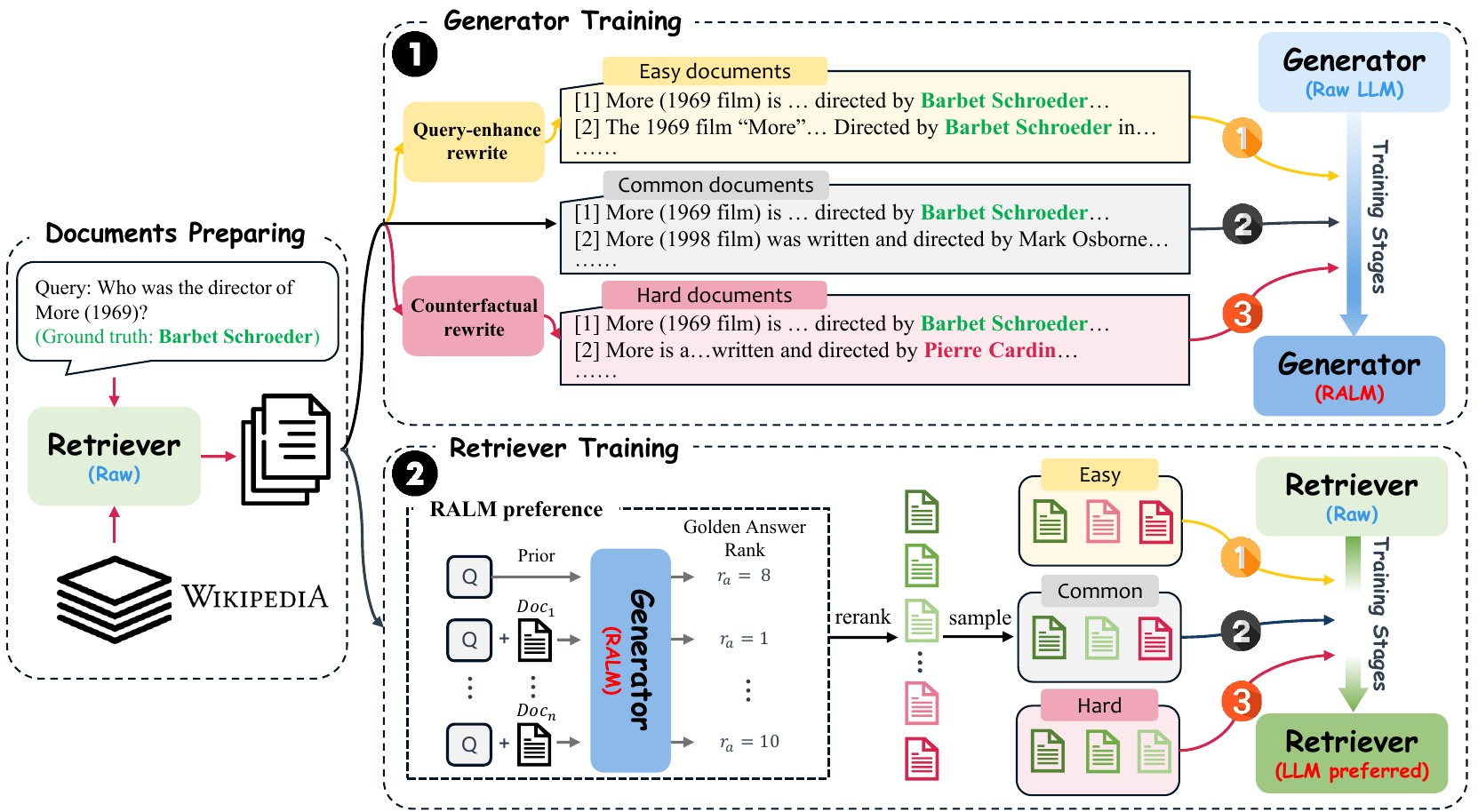}
  \caption {\textbf{The overview of our DACL-RAG training framework,} which has into two continuous phases: (1) Training Generator: We construct multiple difficulty levels of documents and then finetune a RALM in a stage-by-stage manner. (2) Training Retriever: We use the well-trained RALM to assess documents and rerank them. We then construct document data from easy to difficult, and finetune the retriever in a stage-by-stage manner.}
  \label{fig:main}
\end{figure*}

\section{Methodology}
In this section, we will provide a detailed introduction to our DACL-RAG training framework. We first briefly introduce the RAG pipeline in Section~\ref{sec:3.1}. Then, in Sections~\ref{sec:3.2} and~\ref{sec:3.3}, we present the data augmentation and curriculum learning methods we propose for RALM training and retriever training, respectively. An overview of our DACL-RAG training framework is given in Figure~\ref{fig:main}.
\subsection{Preliminary}
\label{sec:3.1}
The RAG system combines a \textbf{Retriever} that retrieves query-relevant documents from external data bases with a \textbf{Generator} that synthesizes responses from retrieved documents.
\paragraph{Retriever}
Given a query $q$, the Retriever aims to retrieve documents $\{d_1,d_2,...,d_n\}$ relevant to the query from an external knowledge base $\mathcal{D}$. In this work, we employ a dense retriever. Specifically, we use a dual encoder to encode the input query $q$ and the documents $d$ into $E(q)$ and $E(d)$. The similarity between them is defined by cosine similarity, which serves as the score for document retrieval:
\begin{equation}
  \label{eq:docscore}
  score_i = cos(E(q),E(d_i)).
\end{equation}
Typically, we select the $top$-$k$ documents with the highest scores as the input to the generator.
\paragraph{Generator}
Given the $top$-$k$ retrieved documents, the goal of the Generator (i.e., the LLM) is to utilize these external documents to better answer the question. Generally, the retrieved documents are concatenated with the query $q$ as contextual information, and then fed into the Generator to produce the answer:
\begin{equation}
  \label{eq:llminput}
  Output = LLM(d_1\oplus d_2\oplus...\oplus d_k,q).
\end{equation}

In this paper, we first employ a raw retriever to retrieve $n \,(n \gg k)$ relevant documents for each query. For generator (LLM) training, we construct curriculum-ordered data from easy to difficult examples by applying data evolution and document rewriting to $top$-$k$ retrieved documents. For retriever training, we leverage the well-trained generator (RALM) to assess the quality of $n$ retrieved documents. We then perform reranking and sampling to construct curriculum-ordered training data to train the retriever to align with RALM's preference.
\subsection{Data Augmentation and Curriculum Learning for RALM}
\label{sec:3.2}
\begin{figure}[t]
  \centering
  \includegraphics[width=\linewidth]{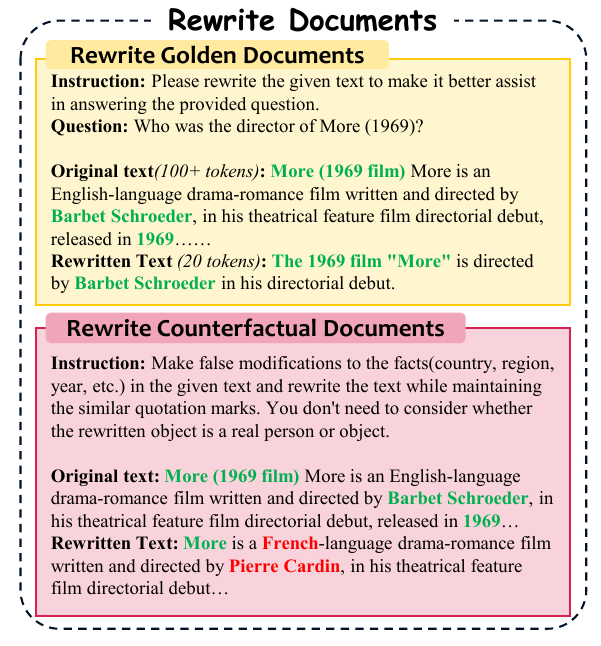}
  \caption {\textbf{Case of document rewriting, including query-enhanced document rewriting and counterfactual document rewriting.} The highlighted \textcolor{mygreen}{\textbf{green}} text represents the keywords within the document that aid in addressing the question, while the highlighted \textcolor{red}{\textbf{red}} text signifies wrong knowledge.}
  \label{fig:prompts}
\end{figure}

To improve the model’s robustness in answering queries with noisy documents, we define the difficulty level of training sets according to the ratio of noise in the documents. Previous work~\cite{yoranmaking, Fang2024EnhancingNR} has considered data augmentation and constructed documents with \textit{relevant noise}, \textit{irrelevant noise}, and \textit{counterfactual noise}. However, they neglected to include simple documents that definitely contain the correct answers, which is essential during the initial training stage for the model to learn the fundamental ability to answer questions from documents that contain the answers.

We first define three levels of difficulty for documents: Easy, Common, and Hard. The Easy level contains at least two documents with correct answers to ensure the model can acquire preliminary answer extraction capabilities. The Common level follows the original retrieved documents, which may not contain the exact correct answers, thereby training the model's ability to infer answers using documents of average quality. The Hard level includes the most challenging documents for the model, potentially containing irrelevant or even harmful noise, which trains the model's ability to counteract disruptive information.

Since the retriever's results may not always include the correct answers, to construct the documents for Easy level, we use the documents provided for the corresponding questions in the MRQA reading comprehension task~\cite{fisch2019mrqa} as the golden documents to ensure that they contain the exact correct answers. Meanwhile, we rewrite the golden document as shown in the upper portion of Figure~\ref{fig:prompts} to obtain document that better meet the model's needs. Our rewriting approach preserves all key information necessary for answering questions while eliminating redundant segments from the original golden documents. Compared to using only the golden document, using both the golden document and its rewritten version ensures that there are at least two documents containing answers within the $top$-$k$ documents, which helps train the model's ability to extract consistent information from multiple documents. The improvements brought by this approach are empirically validated through ablation studies in section \ref{sec:Ablation Study}. We then complete the total number of documents to $k$ using the original retrieved documents.

For Common level, we directly use the $top$-$k$ documents from the retriever's results. Unlike documents in Easy level, these may not contain the correct answers and have more noise related to the questions, which may increase the difficulty for the model to answer the questions.

For Hard level, we randomly select a document from the $top$-$k$ and replace it with retrieved document from other questions or perform counterfactual rewriting as shown in the lower portion of Figure~\ref{fig:prompts}. This introduces irrelevant and counterfactual noise, which represents the most challenging type of documents for the model, aiming to train the model's robustness against various types of real-world data.

To demonstrate the difficulty differences across different levels, we test each level of data using a naive RAG system (with Contriever-MSMARCO as the retriever and LLaMA3-8B-Instruct as the generator). The results are shown in Table~\ref{tab:difficulty_metric_generator}. We established the document difficulty assessment framework based on two perspectives and four key metrics: (1) For LLM generation performance, we evaluate using Exact Match (EM) and F1 scores. (2) For inherent document quality, we evaluate using the recall rate (R@k) of documents (here we use both R@1 and R@5). The definitions of these metrics are detailed in Section~\ref{section: metrics}. The difficulty assessment results are shown in Table~\ref{tab:difficulty_metric_generator}, where we observe significant differences in difficulty levels across various training stages.

During the training phase, data from the three difficulty levels are fed into the model in sequence to conduct Supervised Fine-Tuning (SFT). Through the curriculum learning approach that progresses from easy to difficult, the model is gradually trained to extract answers from documents and to resist the interference of distracting documents.
\subsection{Data Augmentation and Curriculum Learning for Retriever}
\label{sec:3.3}
\begin{table}[!t]
\centering
\caption{Evaluation of Document Data Difficulty Across Different Generator Training Stages}
\label{tab:difficulty_metric_generator}
\begin{tabular}{cccccc}
\toprule
Difficulty & Stages &  EM   &  F1   &  R1   &   R5   \\
\midrule
\rowcolor[HTML]{FFFBDE}
Easy    &   1    & 61.23 & 74.33 & 65.38 & 100.00 \\
\rowcolor[HTML]{EEEEEE}
Common   &   2    & 47.87 & 59.77 & 38.00 & 59.78  \\
\rowcolor[HTML]{FFE1E0}
Hard    &   3    & 41.67 & 52.99 & 20.17 & 54.68  \\
\bottomrule
\end{tabular}
\end{table}

In this section, we will introduce our retriever training strategy. It initially employs a well-trained RALM to rerank retrieved documents and, via stratified sampling, construct three difficulty levels: easy, common, and hard, where simple corresponds to larger inter-document quality gaps for the retriever, and hard corresponds to smaller gaps. The retriever is then trained in stages, following a curriculum learning regime that progressively increases difficulty and aligns the retriever with the RALM.

We first employ the trained RALM to assess documents, thereby eliciting its preferences for them. Differs from that in~\cite{linra}, which relies on the raw LLM, utilizing a well-trained RALM for evaluation provides a more effective means of differentiating between high-quality and low-quality documents.

Previous researchers~\cite{shi2023replug, zhang2024multi} assess document quality based on the probability of the model generating the correct answer and the improvement in decoding rank, respectively. However, using either method alone has its limitation: 
the probability of generating the correct answer can be unstable when the input documents are diverse, 
i.e., there may exist multiple documents containing answers, but due to their diverse content, the probability of generating accurate answers varies significantly, making it difficult to reliably assess document quality.
On the other hand, relying solely on the decoding rank cannot finely distinguish between good and bad documents when the model itself is capable of generating the correct answer, i.e., when most of the input documents have a decoding rank of 1 for the correct answer, indicating gains from the model’s inherent capability rather than document quality.

To address these limitations, we propose using the improvement in decoding rank as a coarse measure of document quality, and then using the probability of generating the correct answer to finely rank documents with the same decoding rank improvement. This helps us obtain a reranked list of retrieved documents:
\begin{equation}
  \label{eq:rerankdoc}
  \mathcal{D}_{rerank} = \{d_1,d_2,...,d_n\}.
\end{equation}

Here, $d_1$ represents the document most beneficial for the LLM to generate the correct answer, i.e., the document that yields both the maximum rank improvement and the highest probability gain for the correct answer when incorporated with question.

Next, we construct data with different difficulty levels for curriculum learning, aiming to first train the retriever to distinguish documents with obvious quality gaps and then train it to distinguish documents with slightly quality differences. Specifically, we divide $\mathcal{D}_{rerank}$ into three groups $\{\mathcal{D}_1,\mathcal{D}_2,\mathcal{D}_3\}$ based on the ranking order from RALM. These groups contain documents in the ranges $[d_1,d_{n_1}]$, $(d_{n_1},d_{n_2}]$, and $(d_{n_2},d_n]$, respectively, representing good documents, sub-optimal documents, and hard negative documents for answering questions.

For the retriever, we expect it to distinguish the difference between good and bad documents, and to assign higher scores to good documents and lower scores to bad documents.
When the document quality in the training data varies significantly (with obvious good and bad documents), the retriever can more easily distinguish the quality differences among documents and assign scores with larger gaps, making training easier.
When the document quality in the training data becomes more consistent (either all relatively good or all relatively poor), the retriever struggles to distinguish between these documents, assigns very similar scores to them, and training becomes more difficult.
Therefore, we define training data with a larger ranking span between different documents in the sampled data as easy data—i.e., the simplest data is sampled across $[d_1,d_{n_1}]$, $(d_{n_1},d_{n_2}]$, and $(d_{n_2},d_n]$. The smaller the ranking span, the more difficult the data is for the retriever—e.g., data sampled only within $[d_1,d_{n_1}]$ is the most difficult.
\begin{table}[!t]
\centering
\caption{Divergence in LLM-based Document Ranking Across Retriever Training Stages
}
\label{tab:doc_rank}
\begin{tabular}{cccc}
\toprule
Difficulty & Stages & Avg. $\Delta$rank & Max $\Delta$rank\\
\midrule
\rowcolor[HTML]{FFFBDE}
Easy    & 1 & 8.39 & 19\\
\rowcolor[HTML]{EEEEEE}
Common & 2 & 4.90 & 14\\
\rowcolor[HTML]{FFE1E0}
Hard   & 3 & 2.00 & 4\\
\bottomrule
\end{tabular}
\end{table}

To construct our stage-wise training data, we sample $k_1$, $k_2$, and $k_3$ documents from each of the three groups, where $k_1+k_2+k_3=k$. Across stages, we increase $k_1$ while decreasing $k_2$ and $k_3$, thereby narrowing the quality gap between the sampled documents.
Table \ref{tab:doc_rank} presents the average ranking divergence of documents across different training phases. Larger values indicate more distinct document rankings, demonstrating that these documents can be more easily distinguished by the retriever. Conversely, smaller ranking differences imply that the documents are more difficult to differentiate in terms of quality.
We have also designed a tiered loss function to better fit our approach: 
\begin{equation}
  \label{eq:lossfunc}
  \mathcal{L}(q,\mathcal{D}_k) = \\
  -\sum_{d_{i,j}\in{\mathcal{D}_k}} \sum_{j>i} \frac{j-i}{n-1} log(\frac{e^{s_i}}{e^{s_i}+e^{s_j}}),
\end{equation}
where $\mathcal{D}_k$ denotes a subset of $k$ documents selected from $\mathcal{D}$, $n$ represents the total number of documents involved in the ranking, and $i$,$j$ are the ranks of the documents, $s_i$ denotes the similarity score between the query $q$ and the document $d_i$. This loss function implies that: (1) the difference in similarity scores should increase with ranking distance, and (2) higher-ranked documents must maintain larger similarity scores. We start by distinguishing documents with obvious quality differences and gradually move to distinguishing documents with slight quality differences. This stage-by-stage approach helps bridge the gap between the original retriever and the retriever aligned with RALM preferences.

\section{Experimental Setting}
\subsection{Datasets}
\label{app:datasets}
\begin{table}[t]
  \centering
  \caption{Details of the datasets we used.}
  \small{
  \begin{tabular}{lcccc}
    \toprule
    \multirow{2}{*}{\textbf{Dataset}} & \multicolumn{3}{c}{\textbf{Train}} & \textbf{Test}\\
    \cmidrule{2-5}
    & Total & RALM & Retriever & Total\\
    \midrule
    \textbf{NQ} & 87372 & 1500 & 30000 & 2837\\
    \textbf{TriviaQA} & 61844 & 1500 & 30000 & 5359\\
    \textbf{PopQA} & 10000 & 1500 & 10000 & 4267\\
    \textbf{HotpotQA} & 88869 & 1500 & 30000 & 5600\\
    \textbf{2Wiki} & - & - & - & 12576\\
    \bottomrule
  \end{tabular}
  }
\label{datasets}
\end{table}

We first evaluated our complete training framework under the standard open-domain Question-Answering (QA) setting. Subsequently, we assessed the robustness of the RALM within our training pipeline under a setting with added noise.
\subsubsection{Open-domain QA}We considered five open-domain QA datasets: single-hop question datasets \textbf{Natural Questions (NQ)}~\cite{10.1162/tacl_a_00276}, \textbf{TriviaQA}~\cite{joshi-etal-2017-triviaqa}, and \textbf{PopQA}~\cite{mallen-etal-2023-trust}, as well as the multi-hop question dataset \textbf{HotpotQA}~\cite{yang-etal-2018-hotpotqa} and \textbf{2WikiMultiHopQA (2Wiki)}\cite{ho-etal-2020-constructing}. For each dataset, we employed Contriever-MSMARCO~\cite{izacardunsupervised} as the retriever. Following previous settings~\cite{asai2024selfrag}, for NQ, TriviaQA, HotpotQA, and 2Wiki, we retrieved the corresponding documents from the 2018 Wikipedia corpus provided by~\cite{izacard2023atlas}. For PopQA, we retrieved the documents from the 2020 Wikipedia corpus. Noting that PopQA only contains a test set, which we split into two non-overlapping parts for training and testing, respectively. We utilized datasets from the KILT benchmark~\cite{petroni-etal-2021-kilt} and open-source platform Hugging Face\footnote{The datasets for NQ, TriviaQA, and HotpotQA were sourced from the KILT benchmark, while PopQA and 2Wiki datasets were obtained from the authors' publicly released data.}. We sampled training data from the training sets of NQ, TriviaQA, HotpotQA, and PopQA, while reserving 2Wiki exclusively for testing purposes without including it in the training process. 
Table~\ref{datasets} details the composition of our training data, where \textbf{RALM} and \textbf{Retriever} denote the randomly sampled subsets from the training set for generator and retriever training, respectively.

\subsubsection{Robustness Test}To evaluate the robustness of the RALM in our framework, we artificially introduced irrelevant and counterfactual documents into the retrieved document sets. Specifically, for each question and its retrieved documents in the test set, we generated test data with irrelevant noise by randomly replacing one of the retrieved documents with a document retrieved for a different question. Additionally, we created test data with counterfactual noise by randomly rewriting one of the retrieved documents in a counterfactual manner. The test sets obtained through these two methods will be used separately to assess the model's robustness against irrelevant and counterfactual noise.
\subsection{Evaluation Metrics}
\label{section: metrics}
We employ \textbf{Exact Match (EM)} and \textbf{F1} score to evaluate the effectiveness of final answer generation. Specifically, EM assesses whether the LLM-generated answers are identical to the correct answers. Meanwhile, the F1 score integrates Precision and Recall to measure the accuracy and coverage of the LLM in generating answers.

For assessing the quality of documents retrieved by the retriever, we employ \textbf{Recall@k (R@k)} as the evaluation metric. This metric indicates the proportion of top-k retrieved documents that contain the correct answer. Specifically, $R@5 = 50\%$ means that for $50\%$ of the questions, their correct answers are included within the top-5 retrieved documents.
\subsection{Baselines}
We utilized several training methods for RALM and Retriever as baselines, and firstly evaluated the performance of all the RALM training methods:
(1) \textbf{RAAT}\cite{Fang2024EnhancingNR}: Enhance the robustness of RALMs through adversarial training.
(2) \textbf{RALM$_{top5}$}: The most basic approach in RALM training by directly employing the $top$-$5$ retrieved documents.
(3) \textbf{RALM$_{golden}$}\cite{linra}: Add the golden document to the retrieved documents to train the RALM.
(4) \textbf{RetRobust}\cite{yoranmaking}: Randomly inject irrelevant and counterfactual documents into the retrieved documents to expose the model to diverse types of noise during training.

Then, we continued to assess the performance of retrievers on the best-performing RALM:
(1) \textbf{Replug}\cite{shi2023replug}: Minimize the KL divergence between the score of retriever's and the model's preference distribution to train a retriever that better aligns with the LLM.
(2) \textbf{RA-DIT}\cite{linra}: Combine the method of finetuning the retriever using KL divergence and finetuning the RALM with retrieved documents that include golden documents as context.
(3) \textbf{LLM-Embedder}\cite{zhang2024multi}: Define the quality of document $d$ based on how much it improves the correct answer's ranking in LLM's response and proposes a fine-grained hierarchical loss function for training.
\begin{table}[ht]
  \centering
  \caption{Prompt for LLM inference.}
  \begin{tabular}{l}
    \toprule
    \textbf{Prompt}\\
    \midrule
    \textbf{System Prompt:} Answer the following questions with two to three \\words. Your answer must be formatted as follows: \\Answer: $<$your answer$>$\\
    \textbf{User Prompt:} The following contexts will help you answer the\\question.\\
    \{paragraph\}\\
    Question: \{instruction\}\\
    \bottomrule
  \end{tabular}
\label{Inference Prompt}
\end{table}
\begin{table*}[!ht]
  \centering
  \caption{\label{main-result}
    \textbf{Experimental results for EM and F1 scores(\%)} on four open-domain QA datasets compared with multiple baselines in RALM or Retriever training. The \underline{underlined} and \textbf{boldface} items indicate the best results under each setting. "$RALM_{CL}$" refers to the RALM trained with our DACL strategy.
  }
  { \small
  \begin{tabular}{lcccccccccc}
    \toprule
    \multirow{2.5}{*}{\textbf{Method}} & \multicolumn{2}{c}{\textbf{NQ}} & \multicolumn{2}{c}{\textbf{TriviaQA}} & \multicolumn{2}{c}{\textbf{PopQA}} & \multicolumn{2}{c}{\textbf{HotpotQA}} & \multicolumn{2}{c}{\textbf{Avg.}} \\
    \cmidrule(lr){2-3}\cmidrule(lr){4-5}\cmidrule(lr){6-7}\cmidrule(lr){8-9}\cmidrule(lr){10-11}
    & EM$\uparrow$ & F1$\uparrow$ & EM$\uparrow$ & F1$\uparrow$ & EM$\uparrow$ & F1$\uparrow$ & EM$\uparrow$ & F1$\uparrow$ & EM$\uparrow$ & F1$\uparrow$\\
    \hline
    \addlinespace[0.2em]

    \multicolumn{11}{l}{$Llama3_{8B}$}\\
    \addlinespace[0.2em]

    \hline
    \hspace{0.8em}$naive$-$RAG$ & 39.51 & 53.37 & 72.74 & 81.80 & 46.07 & 52.16 & 25.23 & 39.46 & 45.89 & 56.70\\
    \hline
    \addlinespace[0.2em]
    \multicolumn{11}{l}{\hspace{0.8em}$With$ $RALM$ $Training$}\\
    \addlinespace[0.2em]
    \hline
    \hspace{0.8em}$RAAT$\cite{Fang2024EnhancingNR} & 31.65 & 40.39 & 71.84 & 77.02 & 38.49 & 41.87 & 22.16 & 30.26 & 41.04 & 47.39\\
    \rowcolor[HTML]{FBFBFB}
    \hspace{0.8em}$RALM_{golden}$\cite{linra} & 42.90 & 52.30 & 79.55 & 83.84 & 50.06 & 52.45 & 31.25 & 41.96 & 50.94 & 57.64\\
    \rowcolor[HTML]{E8F9FF}
    \hspace{0.8em}$RALM_{top5}$\cite{10.5555/3495724.3496517} & 48.15 & 58.95 & 80.32 & 85.05 & 57.58 & 60.13 & 36.57 & 48.46 & 55.66 & 63.15\\
    \rowcolor[HTML]{C4D9FF}
    \hspace{0.8em}$RetRobust$\cite{yoranmaking} & 47.83 & 58.60 & 81.25 & 85.96 & 57.51 & 60.31 & 37.02 & 48.99 & 55.90 & 63.47\\
    \rowcolor[HTML]{C5BAFF}
    \hspace{0.8em}$RALM_{CL}(Ours)$ & \underline{51.53} & \underline{61.19} & \underline{82.93} & \underline{87.59} & \underline{59.71} & \underline{62.28} & \underline{38.14} & \underline{50.22} & \underline{58.08} & \underline{65.32}\\
    \hline
    \addlinespace[0.2em]
    \multicolumn{11}{l}{\hspace{0.8em}$With$ $Retriever$ $Training$}\\
    \addlinespace[0.2em]
    \hline
    \rowcolor[HTML]{FBFBFB}
    \hspace{0.8em}$RA$-$DIT$\cite{linra} & 49.52 & 59.94 & 76.62 & 81.10 & 61.73 & 64.47 & 36.14 & 47.82 & 56.00 & 63.33\\
    \rowcolor[HTML]{E8F9FF}
    \hspace{0.8em}$Replug$\cite{shi2023replug} & 52.52 & 62.28 & 76.86 & 81.59 & 63.75 & 66.35 & 37.75 & 49.22 & 57.72 & 64.86\\
    \rowcolor[HTML]{C4D9FF}
    \hspace{0.8em}$LLM$-$Embedder$\cite{zhang2024multi} & 51.85 & 61.41 & 82.60 & 87.20 & 64.24 & 66.69 & 38.80 & 50.51 & 59.37 & 66.45\\
    \rowcolor[HTML]{C5BAFF}
    \hspace{0.8em}$DACL$-$RAG(Ours)$ & \textbf{53.01} & \textbf{62.51} & \textbf{82.97} & \textbf{87.57} & \textbf{66.63} & \textbf{69.12} & \textbf{39.11} & \textbf{51.19} & \textbf{60.43} & \textbf{67.60}\\

    \hline
    \addlinespace[0.2em]
    \multicolumn{11}{l}{$Qwen2.5_{7B}$}\\
    \addlinespace[0.2em]
    \hline
    \hspace{0.8em}$naive$-$RAG$ &37.65 &50.93 &74.73 &80.65 &46.22 &51.21 &29.38 &39.69 &47.00 &55.62 \\

    \hline
    \addlinespace[0.2em]
    \multicolumn{11}{l}{\hspace{0.8em}$With$ $RALM$ $Training$}\\
    \addlinespace[0.2em]
    \hline
    \hspace{0.8em}$RAAT$\cite{Fang2024EnhancingNR} &30.73 &39.67 &74.24 &79.85 &37.96 &42.59 &27.94 &35.47 &42.72 &49.40 \\
    \rowcolor[HTML]{FBFBFB}
    \hspace{0.8em}$RALM_{golden}$\cite{linra} &41.06 &50.20 &75.93 &80.78 &48.30 &51.04 &30.71 &41.34 &49.00 &55.84 \\
    \rowcolor[HTML]{E8F9FF}
    \hspace{0.8em}$RALM_{top5}$\cite{10.5555/3495724.3496517} &44.80 &55.06 &75.28 &81.04 &55.38 &58.33 &35.21 &46.41 &52.67 &60.21 \\
    \rowcolor[HTML]{C4D9FF}
    \hspace{0.8em}$RetRobust$\cite{yoranmaking} &45.05 &55.35 &77.09 &82.78 &54.93 &57.66 &35.77 &47.45 &53.21 &60.81 \\
    \rowcolor[HTML]{C5BAFF}
    \hspace{0.8em}$RALM_{CL}(Ours)$ & \underline{48.47} & \underline{58.29} & \underline{79.44} & \underline{84.67} & \underline{55.89} & \underline{59.10} & \underline{37.38} & \underline{49.05} & \underline{55.30} & \underline{62.78}\\
    \hline
    \addlinespace[0.2em]
    \multicolumn{11}{l}{\hspace{0.8em}$With$ $Retriever$ $Training$}\\
    \addlinespace[0.2em]
    \hline
    \rowcolor[HTML]{FBFBFB}
    \hspace{0.8em}$RA$-$DIT$\cite{linra} &44.06 &53.54 &75.89 &80.77 &58.45 &61.18 &31.45 &42.07 &52.46 &59.39\\
    \rowcolor[HTML]{E8F9FF}
    \hspace{0.8em}$Replug$\cite{shi2023replug} &49.67 &59.90 &78.95 &84.34 &62.53 &65.68 &35.32 &47.10 &56.62 &64.26 \\
    \rowcolor[HTML]{C4D9FF}
    \hspace{0.8em}$LLM$-$Embedder$\cite{zhang2024multi} &49.45 &59.22 &79.08 &84.34 &60.53 &63.35 &35.86 &47.61 &56.23 &63.63 \\
    \rowcolor[HTML]{C5BAFF}
    \hspace{0.8em}$DACL$-$RAG(Ours)$ & \textbf{51.11} & \textbf{60.56} & \textbf{80.34} & \textbf{85.48} & \textbf{63.56} & \textbf{66.57} & \textbf{36.48} & \textbf{48.47} & \textbf{57.87} & \textbf{65.27}\\
    \bottomrule
  \end{tabular}
  }
\end{table*}
\subsection{Implementation Details}
\subsubsection{RALM Training}We trained our model based on \textbf{LLaMA3-8B-Instruct}\footnote{\url{https://huggingface.co/meta-llama/Meta-Llama-3-8B-Instruct}}\cite{dubey2024llama} and \textbf{Qwen2.5-7B-Instruct}\footnote{\url{https://huggingface.co/Qwen/Qwen2.5-7B-Instruct}}\cite{qwen2.5}, using five retrieved documents. We randomly selected 6000 training samples, 1500 from each of four datasets, and conducted LoRA finetuning based on \textbf{LLaMA Factory}\footnote{\url{https://github.com/hiyouga/LLaMA-Factory}}\cite{zheng2024llamafactory}, with the LoRA rank set to 8, a learning rate of 1e-4, a gradient accumulation step of 8, and a warmup ratio of 0.1. All experiments were conducted on an A800 80G GPU card.
\subsubsection{Retriever Training} We trained our model based on \textbf{Contriever-MSMARCO}~\cite{izacardunsupervised}, a widely used unsupervised dense retriever with strong cross-dataset transfer and broad domain generality. By randomly sampling from the training sets of four datasets, we constructed a retriever training set comprising 100,000 samples. We set the number of input retrieved documents $k$ to 5, the learning rate of 1e-5, the number of epochs to 3, and the batch size to 64. For our three-stage curriculum learning settings, the number of reranked documents $n$ is set to 20. $n_1=[1,3,5]$ for each training epoch and $n_2$ is set consistently to 15. The sampling parameters for each stage are $k_1=[1,3,5]$, $k_2=[2,2,0]$, and $k_3=[2,0,0]$.

\subsubsection{Inference} During inference, we used the same prompt across all models to ensure fairness, as shown in Table~\ref{Inference Prompt}. In this prompt, {paragraph} represents the retrieved document, and {instruction} represents the query. We extracted the content following "Answer:" in the model's response as the final answer.
\section{Results}
\subsection{Main Results}
The overall results are shown in Table~\ref{main-result}, where we compare our method with baselines across two distinct model families.
\subsubsection{With RALM Training}
Results in the upper half of Table~\ref{main-result} show that our method outperforms prior approaches across all test sets on both model backbones, yielding improvements of 2.18\% in average EM and 1.85\% in average F1 over the best baseline, thereby validating the effectiveness and generalization of our approach.
It is noted that the model trained with the RAAT method performs worse than the baseline model. This is because its training data follows the setting of $\{d_{golden}\oplus d_{noise}\}$. While this setting trains the model to be robust against noise, the inclusion of golden documents in all data deviates from the conventional RAG setup, resulting in poorer performance under the standard setting. Training using documents that contain a golden document can enhance the model's basic performance, i.e., its ability to extract correct answers from documents. However, due to the limitations of the retriever, the retrieved documents may not always contain the necessary information to answer the question. Therefore, training solely for answer extraction is insufficient. It is also necessary to further train the model to infer answers from more complex documents. RetRobust considers a more complex setting by accounting for both irrelevant document noise and counterfactual document noise. However, it neglects the incorporation of golden documents, which are beneficial in the early stages of model training. In contrast, our proposed method considers all types of document combinations, thereby enabling the model to adapt to the diverse documents generated by retrieval. Therefore, it achieves the best performance.
\subsubsection{With Retriever Training}
In the settings that incorporate retriever training, our method also achieves the best average performance. As shown in the lower half of Table~\ref{main-result}, RA-DIT utilizes the model from its experimental setting, namely $RALM_{top5}$, while the other methods all employ the best-performing $RALM_{CL}$. We found that on datasets where LLM itself performs well, such as TriviaQA, the gains from further improving the retriever's performance are minimal. However, for datasets like PopQA, which involve questions about specific entities, enhancing the retriever enables it to more accurately fetch entity-related documents, resulting in a significant boost in performance.
\subsubsection{Robustness Test}
\begin{table}[!t]
  \centering
  \caption{\label{robustness result}
    \textbf{Experimental results on robustness test(\%) over irrelevant and counterfactual documents.} We report the average results on each dataset.
  }
  \small{
  \begin{tabular}{lcccc}
    \toprule
    \multirow{2}{*}{\textbf{Method}} & \multicolumn{2}{c}{\textbf{Irrelevant}} & \multicolumn{2}{c}{\textbf{Counterfactual}}\\
    \addlinespace[0.2em]
    \cline{2-5}
    \addlinespace[0.2em]
    & EM$\uparrow$ & F1$\uparrow$ & EM$\uparrow$ & F1$\uparrow$\\
    \addlinespace[0.2em]
    \hline
    \addlinespace[0.2em]
    $Llama3_{8B}$ & 44.10 & 54.79 & 44.32 & 54.55\\
    $RAAT$ & 39.17 & 45.38 & 39.27 & 45.65\\
    \rowcolor[HTML]{FBFBFB}
    $RALM_{golden}$ & 49.36 & 56.07 & 48.25 & 54.73\\
    \rowcolor[HTML]{E8F9FF}
    $RALM_{top5}$ & 53.48 & 61.05 & 53.63 & 61.13\\
    \rowcolor[HTML]{C4D9FF}
    $RetRobust$ & 53.85 & 61.43 & 54.16 & 61.77\\
    \rowcolor[HTML]{C5BAFF}
    $RALM_{CL}$ & \textbf{56.17} & \textbf{63.42} & \textbf{56.21} & \textbf{63.47}\\
    \bottomrule
  \end{tabular}
  }
\end{table}
To demonstrate the superiority of our RALM in combating noise, we also conducted tests under document settings that included irrelevant noise or counterfactual noise. The results are shown in Table~\ref{robustness result}. Under the influence of document noise, the performance of all models declined to some extent. The impact of irrelevant document noise was comparable to that of counterfactual noise. The experimental results show that our model maintained its leading performance even when different types of noise were added.
\subsection{Generalization Studies}
\label{sec:Generalization experiments}
\begin{figure}[!ht]
  \centering
  \includegraphics[width=\linewidth]{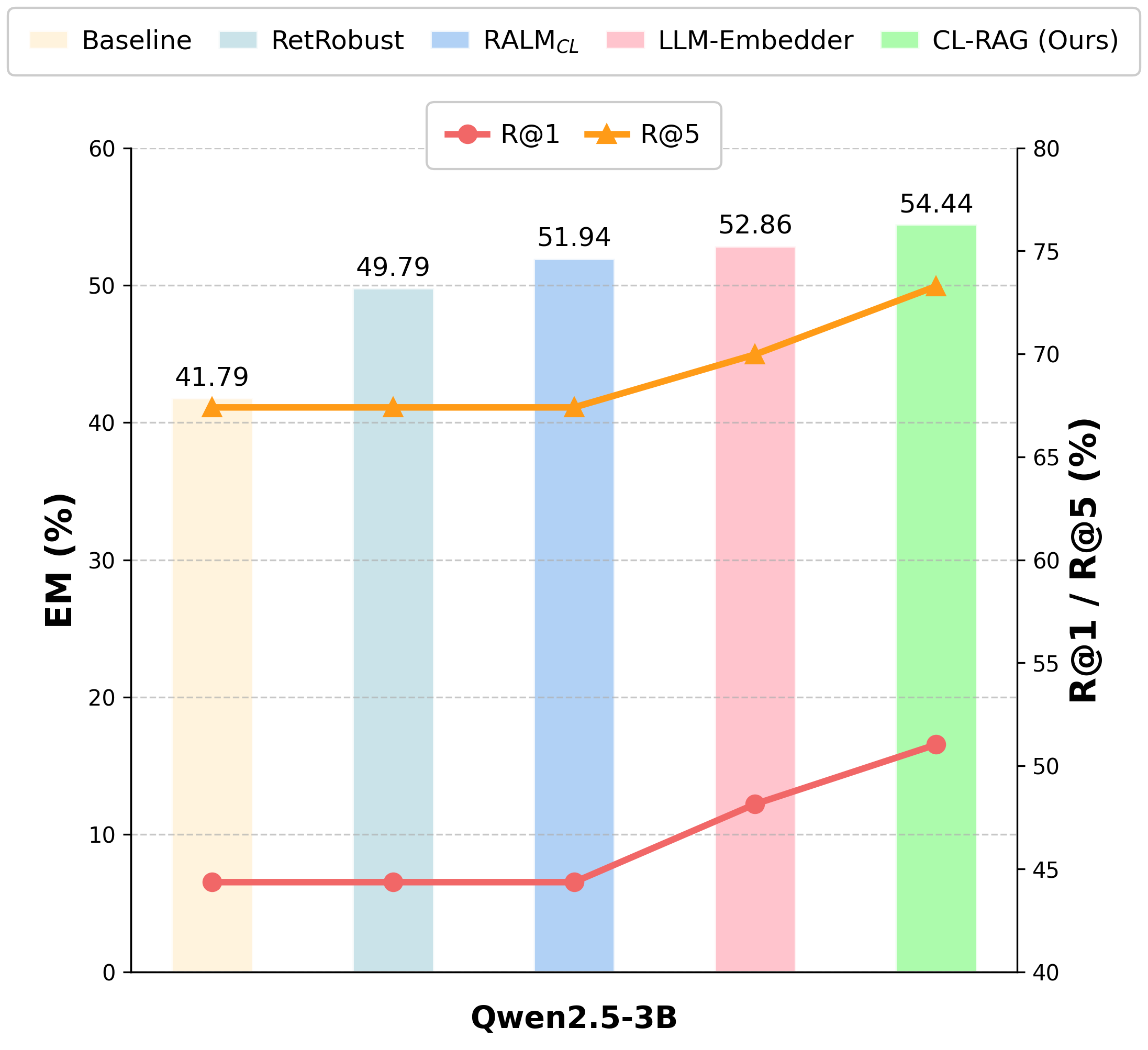}
  \caption {\textbf{Generalization experiments of DACL-RAG training methods across Qwen2.5-3B model,} we report the average Exact Match (EM) scores for generation accuracy evaluation and Recall@k scores for retrieval quality assessment, which demonstrate consistent superiority of our approach.}
  \label{fig:qwen_3B}
\end{figure}
In this section, we will examine the generalization capability of our proposed DACL-RAG training framework, focusing on two key aspects: (1) the generalization of the method when applied to different models, and (2) the generalization of models trained with our method on out-of-domain datasets.
\subsubsection{Generalization experiment of different models}
To evaluate the generalizability of our method across different model parameter scales, we conducted additional experiments using \textbf{Qwen2.5-3B-Instruct}\footnote{\url{https://huggingface.co/Qwen/Qwen2.5-3B-Instruct}} models\cite{qwen2.5}. We tested the performance of generator training as well as the retrieval training guided by this model. The key findings are presented in Figure~\ref{fig:qwen_3B}.

For generator training, We compare our method against the strongest baseline (RetRobust), with all methods using identical documents retrieved by base retriever Contriever-MSMARCO.
The results show that our method still maintains a performance advantage across different model parameters, achieving an average improvement of 10.15\% in EM scores compared to the baseline, and an average performance gain of 2.15\% over the strong baseline RetRobust. This indicates that our method framework generalizes across different model parameters and can be applied to models with varying parameter sizes while maintaining consistent leading performance.

For retriever training, we employ the corresponding well-trained models to score and rank documents, comparing our method against the strong baseline (LLM-Embedder). All generators use corresponding models trained with our DACL-RAG method. Our approach demonstrates superior performance in both document recall rate and final answer accuracy compared to the baselines. These results indicate consistent improvements across different model sizes, confirming the robust generalization capability of our method.
\subsubsection{Generalization experiment of out-of-domain dataset}
\begin{figure}[t]
  \centering
  \includegraphics[width=\linewidth]{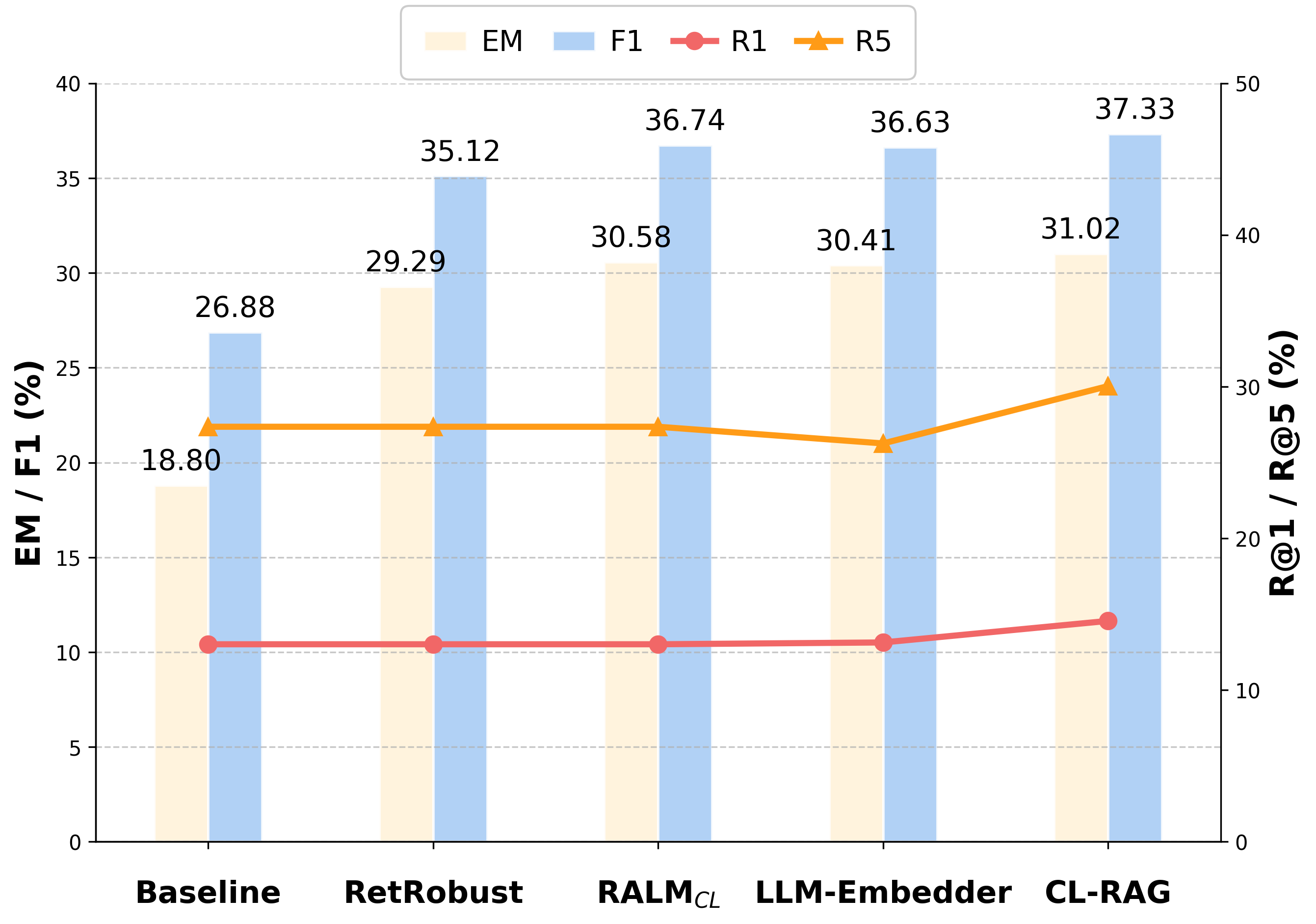}
  \caption {\textbf{Evaluation results on the out-of-domain dataset 2WikiMultiHopQA,} where EM and F1 assess answer accuracy, and R@k evaluates the quality of retrieved documents.}
  \label{fig:wikihopqa}
\end{figure}

To test the generalization ability of our method on unseen data, we conducted evaluations using an out-of-domain dataset. Specifically, we further assessed the performance of our trained retriever and generator on the \textbf{2Wiki}\cite{ho-etal-2020-constructing} dataset (excluded from training), comparing it with strong baseline methods LLM-Embedder and RetRobust. The results are shown in Figure~\ref{fig:wikihopqa}, where EM and F1 scores reflect the performance of the final generated answers, while R@k measures the retriever’s retrieval performance.

The experimental results demonstrate that our method achieves strong performance even on untrained datasets, outperforming the baseline approaches. We observed that in terms of document retrieval recall, while LLM-Embedder shows a slight improvement in R@1, its R@5 performance actually decreases by 1\% compared to the original retriever, which also negatively impacts its final answer generation performance. In contrast, our method demonstrates excellent performance in overall recall rates, improving R@1 by 1.54\% while simultaneously increasing R@5 by 2.68\%, delivering consistent performance improvements. This indicates that our method achieves better results than previous methods in retrieving multiple relevant documents, while also demonstrating the excellent generalization capability of our method.
\subsection{Ablation Studies}
\label{sec:Ablation Study}
In this section, we will discuss each component of our DACL-RAG framework, including: (1) research on document rewriting, (2) in-depth study on retriever training, and (3) ablation study of each training stage. In these experiments, we select the aforementioned four main datasets: NQ, TriviaQA, PopQA, and HotpotQA, using the LLaMA3-8B-Instruct model as the base generator and Contriever-MSMARCO as the base retriever.
\subsubsection{Research on Document Rewriting}
\begin{figure}[t]
  \centering
  \includegraphics[width=\linewidth]{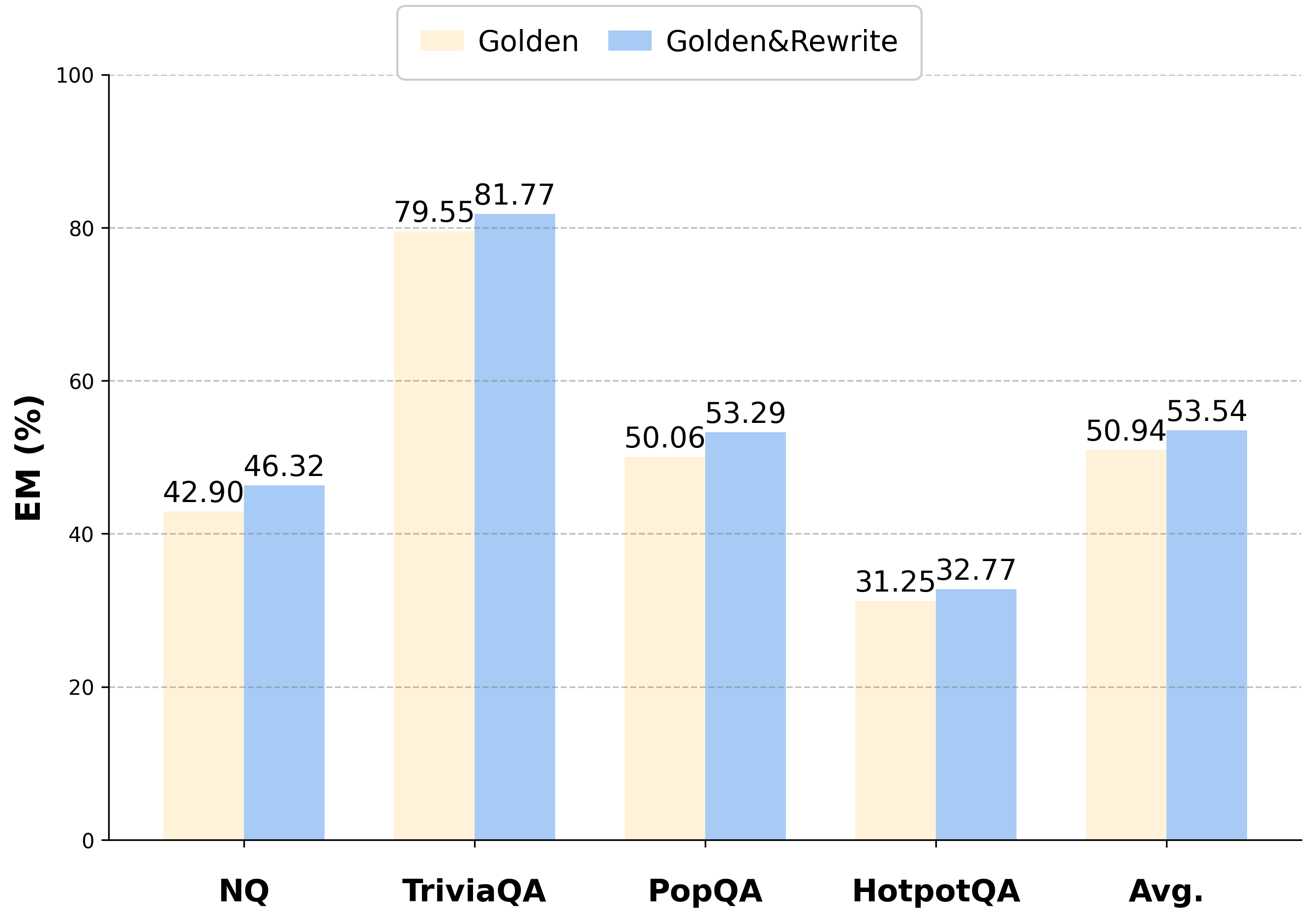}
  \caption {\textbf{Performance comparison between training with golden documents only versus combined golden and rewritten documents.} The joint training approach demonstrates consistent advantages across all test sets.}
  \label{fig:golden&rewrite}
\end{figure}
\begin{figure}[!ht]
  \centering
  \includegraphics[width=\linewidth]{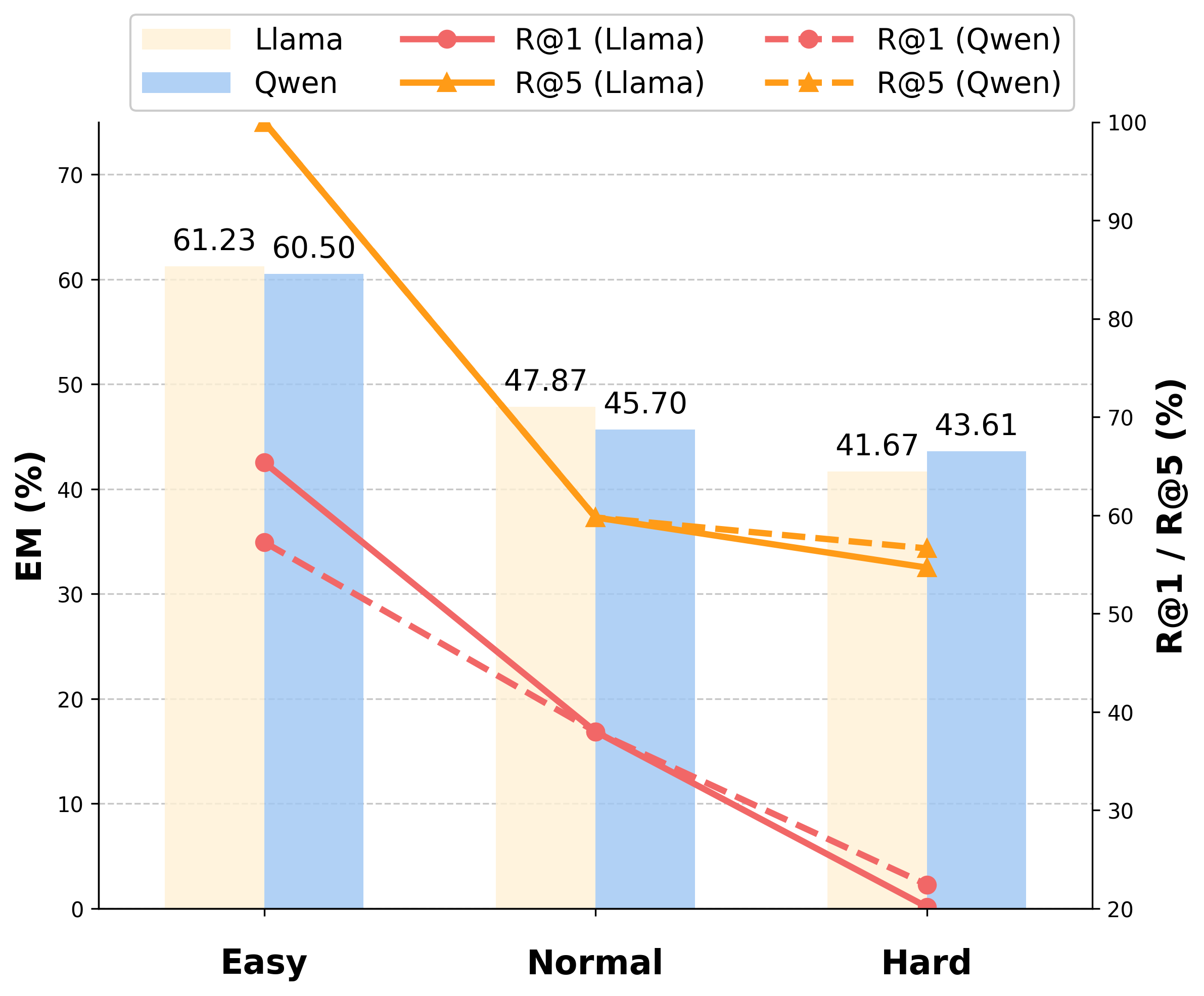}
  \caption {\textbf{Comparative analysis of data quality across different rewrite models at different training stages}, we report average results for each training dataset, revealing quality differences between Llama and Qwen models in document rewriting.}
  \label{fig:difficulty_comparison}
\end{figure}
 For the document rewriting strategy proposed in our method, we conducted further research in two aspects: whether using both rewritten documents and golden documents for training actually enhances model performance compared to using only golden documents, and a study on rewriting quality when using different models.
 
 For the first question, we compared the performance differences when training with only 5 documents containing golden documents versus training with 5 documents containing both rewritten golden documents and original golden documents. The results are shown in Figure \ref{fig:golden&rewrite}. The results show that compared to the method using only golden documents for training, our method has advantages across all datasets, bringing an average improvement of 2.67\% in EM score. This indicates that in data already containing golden documents, adding rewritten documents that also contain correct answers, thereby ensuring the document set contains two or more documents with correct answers, can effectively enable the model to learn the ability to select documents with higher consistency from multiple different documents for answering, thus improving performance.
 
 Regarding rewriting document quality, we compared the rewriting quality based on Llama model and Qwen model (here we used the Qwen2.5-7B-Instruct model), and evaluate the impact of different training stages on LLM generation accuracy and document recall rate. The results are shown in Figure \ref{fig:difficulty_comparison}, where for common difficulty the documents are not rewritten and thus share the same recall rate, while for easy difficulty the top five documents are guaranteed to contain the correct answer, resulting in 100\% R@5. Our analysis reveals that Llama demonstrates superior rewriting performance compared to Qwen across both document types: it achieves higher average recall on first-stage training data and lower average recall on third-stage training data, indicating better performance in both golden document rewriting and fake document rewriting. This suggests Llama may be the more optimal choice for rewriting tasks.
\subsubsection{In-depth Study on Retriever Training}
To demonstrate that using the well-trained RALM to guide the finetuning of the retriever is superior, we compared the retrieval results of the retriever finetuned with the raw LLM preferences to those finetuned with the preferences of RALM. We reported the recall rate of correct answers in the retrieved documents in Table~\ref{retriever study}.

The experimental results show that finetuning with the preferences of RALM generally outperforms finetuning with the raw LLM preferences. We observed that Replug - which evaluates document quality through answer probability - exhibited an unexpected recall rate decrease when using raw LLM, underperforming even the untuned baseline retriever. The instability arises because when the LLM can independently generate correct answers, using answer probability as a document quality metric becomes unreliable. Specifically, in such cases the LLM tends to assign inflated scores to all documents, making it impossible to discern their actual quality differences. In contrast, the well-trained RALM has a higher ability to discern documents and can more accurately differentiate between good and bad ones, thereby achieving better training outcomes. Meanwhile, the results also show that our model demonstrates leading performance across different recall metrics, with more pronounced improvements in R@5 and R@10, indicating that our model has greater advantages when retrieving multiple relevant documents.
\begin{table}[!t]
  \centering
  \caption{\label{retriever study}
    \textbf{Experimental results on Recall rates(\%) for retrievers trained with different methods.} "Baseline" refers to the retrieval results of the raw retriever and "w/o RALM" indicates using the raw LLM without employing the well-trained RALM.
  }
  \small{
  \begin{tabular}{lccc}
    \toprule
    \multirow{2}{*}{\textbf{Method}} & \multicolumn{3}{c}{\textbf{Avg.}}\\
    \addlinespace[0.3em]
    \cline{2-4}
    \addlinespace[0.3em]
    & R@1$\uparrow$ & R@5$\uparrow$ & R@10$\uparrow$ \\
    \addlinespace[0.3em]
    \hline
    \addlinespace[0.3em]
    \rowcolor[HTML]{FBFBFB}
    $Baseline$ & 44.35 & 67.40 & 74.26\\
    \hline
    \rowcolor[HTML]{E8F9FF}
    $Replug$ & 49.10 & 72.32 & 78.43\\
    \rowcolor[HTML]{E8F9FF}
    $w/o\ RALM$ & 43.82 & 67.25 & 74.25\\
    \hline
    \rowcolor[HTML]{C4D9FF}
    $LLM$-$Embedder$ & 49.45 & 71.48 & 77.78\\
    \rowcolor[HTML]{C4D9FF}
    $w/o\ RALM$ & 47.78 & 69.78 & 76.29\\
    \hline
    \rowcolor[HTML]{C5BAFF}
    $DACL$-$RAG$ & \textbf{50.40} & \textbf{73.29} & \textbf{79.55}\\
    \rowcolor[HTML]{C5BAFF}
    $w/o\ RALM$ & 48.55 & 70.71 & 77.33\\
    \bottomrule
  \end{tabular}
  }
\end{table}
\begin{table}[ht]
\centering
\caption{\textbf{Ablation studies for training stages in our DACL-RAG framework(\%).} We report the average EM and F1 scores here.}
\resizebox{\linewidth}{!}{%
\begin{tabular}{cccccccc}
\toprule
\multicolumn{3}{c}{Training Stage} & \multicolumn{2}{c}{Generator Training} & \multicolumn{2}{c}{Retriever Training}\\
\cmidrule(lr){1-3} \cmidrule(lr){4-5} \cmidrule(lr){6-7}
Stage1 & Stage2 & Stage3 & EM$\uparrow$ & F1$\uparrow$ & EM$\uparrow$ & F1$\uparrow$\\
\midrule
$\checkmark$ & & & 53.54 & 60.46 & 59.56 & 66.70\\
& $\checkmark$ & & 55.66 & 63.15 & 59.83 & 67.01\\
& & $\checkmark$ & 55.90 & 63.47 & 58.63 & 65.87\\
\rowcolor[HTML]{FBFBFB}
$\checkmark$ & $\checkmark$ & & 57.56 & 64.72 & 60.04 & 67.17\\
\rowcolor[HTML]{E8F9FF}
& $\checkmark$ & $\checkmark$ & 57.59 & 64.91 & 59.88 & 67.05\\
\rowcolor[HTML]{C4D9FF}
$\checkmark$ & & $\checkmark$ & 57.44 & 64.52 & 60.14 & 67.31\\
\rowcolor[HTML]{C5BAFF}
$\checkmark$ & $\checkmark$ & $\checkmark$ & \textbf{58.08} & \textbf{65.32} & \textbf{60.43} & \textbf{67.60}\\
\bottomrule
\end{tabular}
}
\label{tab:ablation1}
\end{table}
\subsubsection{Ablation Studies of Each Training Stage}
\begin{figure}[!t]
\centering
\includegraphics[width=0.9\linewidth]{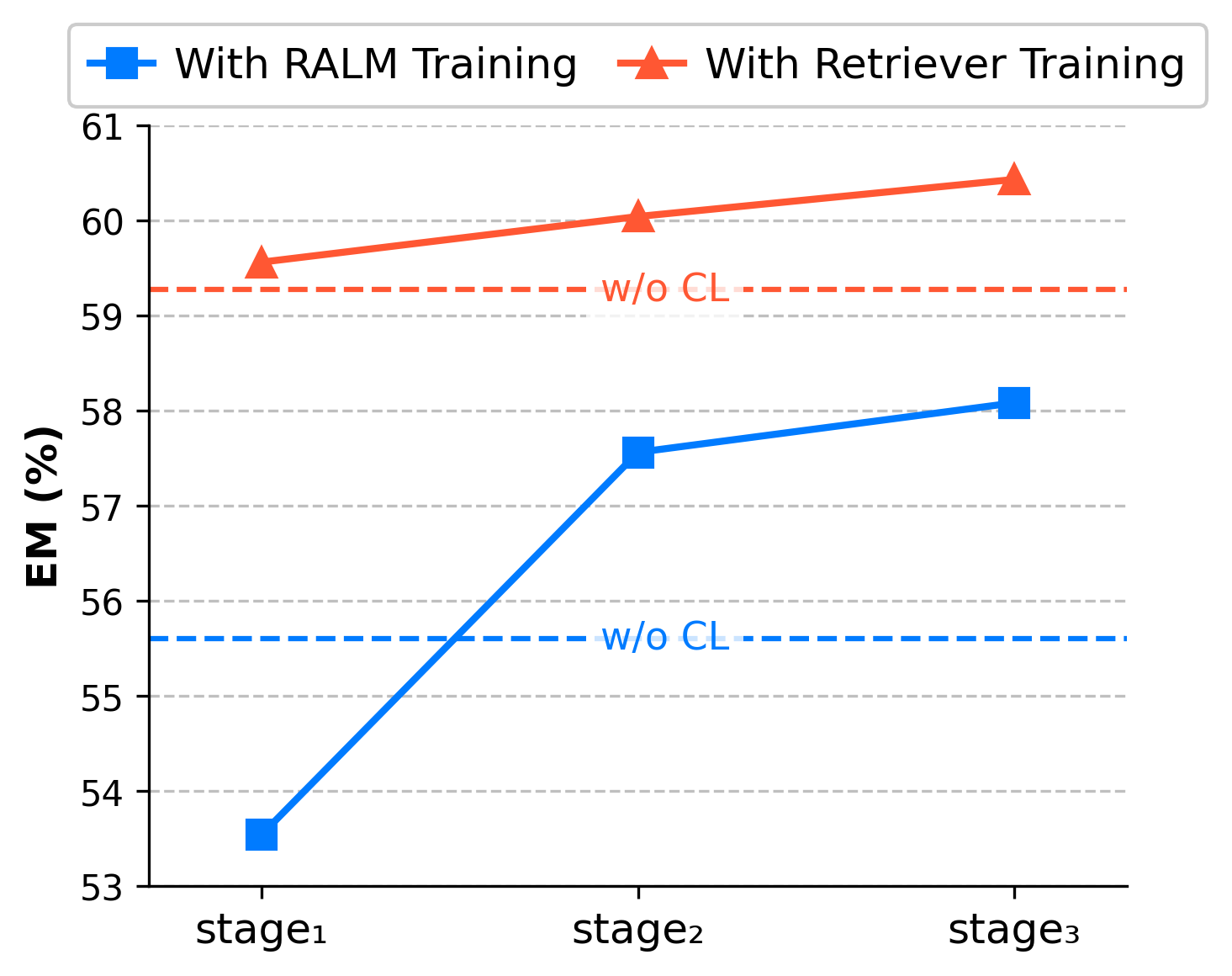}
\caption{\textbf{Ablation studies for each training stage in our DACL-RAG framework.} We demonstrate the EM score gains contributed by each training stage and compare with the results without curriculum learning strategies (w/o CL). "With Retriever Training" employs the well-trained RALM (obtained during generator training) for both retriever training and final answer generation.}
\label{fig:em_ablation}
\end{figure}
To systematically elucidate the contribution of each component to the overall performance, we conducted ablation studies in two aspects: ablation of individual training stages, and ablation of the curriculum learning strategy. 

In the ablation study on training stages, we combined data from each stage and separately tested the performance of training the generator and the retriever. Specifically, we sampled from the training data of easy, common, and hard difficulty levels, keeping the total number of samples constant while varying the difficulty types contained in the samples, ensuring consistency between the samples and the number of backpropagation steps during training, and trained the model in order of difficulty. The results are shown in Table~\ref{tab:ablation1}.

In terms of generator training, we found that the absence of training data from any single stage led to suboptimal performance, and the model trained solely on easy data performed the worst. This indicates that in real retrieval scenarios, where retrieved documents vary in quality, using only easy data fails to train the model to resist interference from complex documents. Although training with common and hard difficulty data achieved relatively good performance, it was still not optimal compared to incorporating easy data, demonstrating the necessity of easy documents in training.

For retriever training, we observed that the model trained exclusively on hard samples performed the poorest. This is likely because the training samples contained only high-quality, less diverse documents, making it difficult for the model to develop strong discriminative ability. Similarly, after training with documents of varying difficulty levels, the model achieved optimal performance.

In parallel, we conducted an ablation study on the curriculum learning strategy. Specifically, we compared the model trained with the curriculum learning strategy (progressing stage by stage) against the model trained by directly feeding samples in random order. We demonstrated the improvement in model capability at each training stage, as well as the test results without using the curriculum learning strategy (w/o CL). The key findings are presented in Figure \ref{fig:em_ablation}, where "with RALM training" indicates training only the generator, and "with retriever training" represents the results obtained by training the retriever on top of the pre-trained generator.

We observed that the performance of both the retriever and the generator steadily improves as the training stages progress. Furthermore, the removal of the curriculum learning strategy resulted in a significant performance decline. Notably, the model trained without curriculum learning for the retriever performed even worse than the model trained with only a single stage, indicating that neglecting the difficulty order of training samples leads to sub-optimal training outcomes.
\subsubsection{Case Study}
Table~\ref{case study of doc} presents the case study comparing the document evaluation result of the raw LLM and the well-trained RALM. The result indicate that for some documents containing implicit knowledge required for answering questions, the raw LLM may generate erroneous judgments, whereas the well-trained RALM is more capable of distinguishing the quality of documents. This also highlights the necessity of the iterative process of training the RALM first and then training the retriever, as the well-trained RALM and the raw LLM may no longer share the same preferences for documents. Separate training may lead to suboptimal final outcomes.
\begin{table}
  \centering
  \caption{\label{case study of doc}
    Case Study of document evaluation using raw LLM and well-trained RALM.
  }
  \begin{tabular}{l}
    \hline
    \textbf{Question}: Who scored a film based on a 1961 science fiction novel\\by Stanislaw Lem?\\
    \textbf{Answer}: Cliff Martinez\\
    \hline
    \textbf{Retrieved Document}\\
    \hline
    \textcolor{mygreen}{\textbf{Stanisław Lem}}\\
    (Poland, Germany, and the Soviet Union). Franz Rottensteiner,\\Lem's former agent abroad, had this to say about Lem's reception \\on international markets: His best-known novels include \textcolor{mygreen}{\textbf{"Solaris"}}\\\textcolor{mygreen}{\textbf{(1961)}}, "His Master's Voice" ("Głos pana", 1968), and the late\\"Fiasco" ("Fiasko", 1987). "Solaris"  was made into a film in 1968\\by Russian director Boris Nirenburg, a film in 1972 by Russian\\director Andrei Tarkovsky—which won a Special Jury Prize at the\\Cannes Film Festival in 1972—andan American film in 2002 by\\Steven Soderbergh. \textcolor{mygreen}{\textbf{"Solaris" is not the only work of Lem's}}\\ \textcolor{mygreen}{\textbf{to be filmed.}}\\
    \hline
    \textbf{Raw LLM preference}\\
    \hline
    Answer probably: \textcolor{red}{0.002}\\
    Answer rank: \textcolor{red}{1417}\\
    \hline
    \textbf{Well-trained RALM preference}\\
    \hline
    Answer probably: \textcolor{mygreen}{0.73}\\
    Answer rank: \textcolor{mygreen}{1}\\
    \hline
  \end{tabular}
\end{table}

\section{Conclusion}
In this study, we introduce DACL-RAG, a multi-level Data Augmentation strategy with a multi-stage Curriculum Learning paradigm for the RAG system. To the best of our knowledge, it's one of the first time that the idea of human imitation learning is integrated with RAG training, which can effectively enhance the generalization and stability of the RAG system. Experiments on five open-domain question answering datasets on three different models provide substantial evidence of the framework's effectiveness and generalization capability. Additionally, separate experiments conducted on the Retriever and the Generator have demonstrated the significant enhancements our method brings to each individual part.


\appendices

\bibliographystyle{IEEEtran}
\bibliography{main}

\begin{thebibliography}{10}
\providecommand{\url}[1]{#1}
\csname url@samestyle\endcsname
\providecommand{\newblock}{\relax}
\providecommand{\bibinfo}[2]{#2}
\providecommand{\BIBentrySTDinterwordspacing}{\spaceskip=0pt\relax}
\providecommand{\BIBentryALTinterwordstretchfactor}{4}
\providecommand{\BIBentryALTinterwordspacing}{\spaceskip=\fontdimen2\font plus
\BIBentryALTinterwordstretchfactor\fontdimen3\font minus
  \fontdimen4\font\relax}
\providecommand{\BIBforeignlanguage}[2]{{%
\expandafter\ifx\csname l@#1\endcsname\relax
\typeout{** WARNING: IEEEtran.bst: No hyphenation pattern has been}%
\typeout{** loaded for the language `#1'. Using the pattern for}%
\typeout{** the default language instead.}%
\else
\language=\csname l@#1\endcsname
\fi
#2}}
\providecommand{\BIBdecl}{\relax}
\BIBdecl

\bibitem{brown2020language}
T.~Brown, B.~Mann, N.~Ryder, M.~Subbiah, J.~D. Kaplan, P.~Dhariwal,
  A.~Neelakantan, P.~Shyam, G.~Sastry, A.~Askell \emph{et~al.}, ``Language
  models are few-shot learners,'' \emph{Advances in neural information
  processing systems}, vol.~33, pp. 1877--1901, 2020.

\bibitem{anil2023palm}
R.~Anil, A.~M. Dai, O.~Firat, M.~Johnson, D.~Lepikhin, A.~Passos, S.~Shakeri,
  E.~Taropa, P.~Bailey, Z.~Chen \emph{et~al.}, ``Palm 2 technical report,''
  \emph{arXiv preprint arXiv:2305.10403}, 2023.

\bibitem{dubey2024llama}
A.~Dubey, A.~Jauhri, A.~Pandey, A.~Kadian, A.~Al-Dahle, A.~Letman, A.~Mathur,
  A.~Schelten, A.~Yang, A.~Fan \emph{et~al.}, ``The llama 3 herd of models,''
  \emph{arXiv preprint arXiv:2407.21783}, 2024.

\bibitem{roberts2020much}
A.~Roberts, C.~Raffel, and N.~Shazeer, ``How much knowledge can you pack into
  the parameters of a language model?'' in \emph{Proceedings of the 2020
  Conference on Empirical Methods in Natural Language Processing (EMNLP)},
  2020, pp. 5418--5426.

\bibitem{kandpal2023large}
N.~Kandpal, H.~Deng, A.~Roberts, E.~Wallace, and C.~Raffel, ``Large language
  models struggle to learn long-tail knowledge,'' in \emph{International
  Conference on Machine Learning}.\hskip 1em plus 0.5em minus 0.4em\relax PMLR,
  2023, pp. 15\,696--15\,707.

\bibitem{Gao2023RetrievalAugmentedGF}
\BIBentryALTinterwordspacing
Y.~Gao, Y.~Xiong, X.~Gao, K.~Jia, J.~Pan, Y.~Bi, Y.~Dai, J.~Sun, Q.~Guo,
  M.~Wang, and H.~Wang, ``Retrieval-augmented generation for large language
  models: A survey,'' \emph{ArXiv}, vol. abs/2312.10997, 2023. [Online].
  Available: \url{https://api.semanticscholar.org/CorpusID:266359151}
\BIBentrySTDinterwordspacing

\bibitem{10.5555/3495724.3496517}
P.~Lewis, E.~Perez, A.~Piktus, F.~Petroni, V.~Karpukhin, N.~Goyal,
  H.~K\"{u}ttler, M.~Lewis, W.-t. Yih, T.~Rockt\"{a}schel, S.~Riedel, and
  D.~Kiela, ``Retrieval-augmented generation for knowledge-intensive nlp
  tasks,'' in \emph{Proceedings of the 34th International Conference on Neural
  Information Processing Systems}, ser. NIPS '20.\hskip 1em plus 0.5em minus
  0.4em\relax Red Hook, NY, USA: Curran Associates Inc., 2020.

\bibitem{min-etal-2020-ambigqa}
\BIBentryALTinterwordspacing
S.~Min, J.~Michael, H.~Hajishirzi, and L.~Zettlemoyer, ``{A}mbig{QA}: Answering
  ambiguous open-domain questions,'' in \emph{Proceedings of the 2020
  Conference on Empirical Methods in Natural Language Processing (EMNLP)},
  B.~Webber, T.~Cohn, Y.~He, and Y.~Liu, Eds.\hskip 1em plus 0.5em minus
  0.4em\relax Online: Association for Computational Linguistics, Nov. 2020, pp.
  5783--5797. [Online]. Available:
  \url{https://aclanthology.org/2020.emnlp-main.466/}
\BIBentrySTDinterwordspacing

\bibitem{lewis-etal-2020-bart}
\BIBentryALTinterwordspacing
M.~Lewis, Y.~Liu, N.~Goyal, M.~Ghazvininejad, A.~Mohamed, O.~Levy, V.~Stoyanov,
  and L.~Zettlemoyer, ``{BART}: Denoising sequence-to-sequence pre-training for
  natural language generation, translation, and comprehension,'' in
  \emph{Proceedings of the 58th Annual Meeting of the Association for
  Computational Linguistics}, D.~Jurafsky, J.~Chai, N.~Schluter, and
  J.~Tetreault, Eds.\hskip 1em plus 0.5em minus 0.4em\relax Online: Association
  for Computational Linguistics, Jul. 2020, pp. 7871--7880. [Online].
  Available: \url{https://aclanthology.org/2020.acl-main.703/}
\BIBentrySTDinterwordspacing

\bibitem{izacard2023atlas}
G.~Izacard, P.~Lewis, M.~Lomeli, L.~Hosseini, F.~Petroni, T.~Schick,
  J.~Dwivedi-Yu, A.~Joulin, S.~Riedel, and E.~Grave, ``Atlas: Few-shot learning
  with retrieval augmented language models,'' \emph{Journal of Machine Learning
  Research}, vol.~24, no. 251, pp. 1--43, 2023.

\bibitem{shi2023replug}
W.~Shi, S.~Min, M.~Yasunaga, M.~Seo, R.~James, M.~Lewis, L.~Zettlemoyer, and
  W.-t. Yih, ``Replug: Retrieval-augmented black-box language models,''
  \emph{arXiv preprint arXiv:2301.12652}, 2023.

\bibitem{yoranmaking}
O.~Yoran, T.~Wolfson, O.~Ram, and J.~Berant, ``Making retrieval-augmented
  language models robust to irrelevant context,'' in \emph{The Twelfth
  International Conference on Learning Representations}, 2023.

\bibitem{linra}
X.~V. Lin, X.~Chen, M.~Chen, W.~Shi, M.~Lomeli, R.~James, P.~Rodriguez,
  J.~Kahn, G.~Szilvasy, M.~Lewis \emph{et~al.}, ``Ra-dit: Retrieval-augmented
  dual instruction tuning,'' in \emph{The Twelfth International Conference on
  Learning Representations}, 2023.

\bibitem{Fang2024EnhancingNR}
\BIBentryALTinterwordspacing
F.~Fang, Y.~Bai, S.~Ni, M.~Yang, X.~Chen, and R.~Xu, ``Enhancing noise
  robustness of retrieval-augmented language models with adaptive adversarial
  training,'' in \emph{Annual Meeting of the Association for Computational
  Linguistics}, 2024. [Online]. Available:
  \url{https://api.semanticscholar.org/CorpusID:270199429}
\BIBentrySTDinterwordspacing

\bibitem{ke-etal-2024-bridging}
\BIBentryALTinterwordspacing
Z.~Ke, W.~Kong, C.~Li, M.~Zhang, Q.~Mei, and M.~Bendersky, ``Bridging the
  preference gap between retrievers and {LLM}s,'' in \emph{Proceedings of the
  62nd Annual Meeting of the Association for Computational Linguistics (Volume
  1: Long Papers)}, L.-W. Ku, A.~Martins, and V.~Srikumar, Eds.\hskip 1em plus
  0.5em minus 0.4em\relax Bangkok, Thailand: Association for Computational
  Linguistics, Aug. 2024, pp. 10\,438--10\,451. [Online]. Available:
  \url{https://aclanthology.org/2024.acl-long.562/}
\BIBentrySTDinterwordspacing

\bibitem{xu2024recomp}
\BIBentryALTinterwordspacing
F.~Xu, W.~Shi, and E.~Choi, ``{RECOMP}: Improving retrieval-augmented {LM}s
  with context compression and selective augmentation,'' in \emph{The Twelfth
  International Conference on Learning Representations}, 2024. [Online].
  Available: \url{https://openreview.net/forum?id=mlJLVigNHp}
\BIBentrySTDinterwordspacing

\bibitem{huang2024raven}
\BIBentryALTinterwordspacing
J.~Huang, W.~Ping, P.~Xu, M.~Shoeybi, K.~Chang, and B.~Catanzaro, ``{RAVEN}:
  In-context learning with retrieval-augmented encoder-decoder language
  models,'' in \emph{First Conference on Language Modeling}, 2024. [Online].
  Available: \url{https://openreview.net/forum?id=GMalvQu0XL}
\BIBentrySTDinterwordspacing

\bibitem{wang2024unims}
H.~Wang, W.~Huang, Y.~Deng, R.~Wang, Z.~Wang, Y.~Wang, F.~Mi, J.~Z. Pan, and
  K.-F. Wong, ``Unims-rag: A unified multi-source retrieval-augmented
  generation for personalized dialogue systems,'' \emph{arXiv preprint
  arXiv:2401.13256}, 2024.

\bibitem{liu2024chatqa}
Z.~Liu, W.~Ping, R.~Roy, P.~Xu, C.~Lee, M.~Shoeybi, and B.~Catanzaro, ``Chatqa:
  Surpassing gpt-4 on conversational qa and rag,'' \emph{Advances in Neural
  Information Processing Systems}, vol.~37, pp. 15\,416--15\,459, 2024.

\bibitem{wang2024retrieval}
M.~Wang, I.~Shafran, H.~Soltau, W.~Han, Y.~Cao, D.~Yu, and L.~El~Shafey,
  ``Retrieval augmented end-to-end spoken dialog models,'' in \emph{ICASSP
  2024-2024 IEEE International Conference on Acoustics, Speech and Signal
  Processing (ICASSP)}.\hskip 1em plus 0.5em minus 0.4em\relax IEEE, 2024, pp.
  12\,056--12\,060.

\bibitem{zhang2024multi}
P.~Zhang, Z.~Liu, S.~Xiao, Z.~Dou, and J.-Y. Nie, ``A multi-task embedder for
  retrieval augmented llms,'' in \emph{Proceedings of the 62nd Annual Meeting
  of the Association for Computational Linguistics (Volume 1: Long Papers)},
  2024, pp. 3537--3553.

\bibitem{zhang-etal-2024-arl2}
\BIBentryALTinterwordspacing
L.~Zhang, Y.~Yu, K.~Wang, and C.~Zhang, ``{ARL}2: Aligning retrievers with
  black-box large language models via self-guided adaptive relevance
  labeling,'' in \emph{Proceedings of the 62nd Annual Meeting of the
  Association for Computational Linguistics (Volume 1: Long Papers)}, L.-W. Ku,
  A.~Martins, and V.~Srikumar, Eds.\hskip 1em plus 0.5em minus 0.4em\relax
  Bangkok, Thailand: Association for Computational Linguistics, Aug. 2024, pp.
  3708--3719. [Online]. Available:
  \url{https://aclanthology.org/2024.acl-long.203/}
\BIBentrySTDinterwordspacing

\bibitem{izacard2020leveraging}
G.~Izacard and E.~Grave, ``Leveraging passage retrieval with generative models
  for open domain question answering,'' \emph{arXiv preprint arXiv:2007.01282},
  2020.

\bibitem{NEURIPS2024_db93ccb6}
\BIBentryALTinterwordspacing
Y.~Yu, W.~Ping, Z.~Liu, B.~Wang, J.~You, C.~Zhang, M.~Shoeybi, and
  B.~Catanzaro, ``Rankrag: Unifying context ranking with retrieval-augmented
  generation in llms,'' in \emph{Advances in Neural Information Processing
  Systems}, A.~Globerson, L.~Mackey, D.~Belgrave, A.~Fan, U.~Paquet,
  J.~Tomczak, and C.~Zhang, Eds., vol.~37.\hskip 1em plus 0.5em minus
  0.4em\relax Curran Associates, Inc., 2024, pp. 121\,156--121\,184. [Online].
  Available:
  \url{https://proceedings.neurips.cc/paper_files/paper/2024/file/db93ccb6cf392f352570dd5af0a223d3-Paper-Conference.pdf}
\BIBentrySTDinterwordspacing

\bibitem{li2025ragddroptimizingretrievalaugmentedgeneration}
\BIBentryALTinterwordspacing
X.~Li, S.~Mei, Z.~Liu, Y.~Yan, S.~Wang, S.~Yu, Z.~Zeng, H.~Chen, G.~Yu, Z.~Liu,
  M.~Sun, and C.~Xiong, ``Rag-ddr: Optimizing retrieval-augmented generation
  using differentiable data rewards,'' 2025. [Online]. Available:
  \url{https://arxiv.org/abs/2410.13509}
\BIBentrySTDinterwordspacing

\bibitem{bengio2009curriculum}
Y.~Bengio, J.~Louradour, R.~Collobert, and J.~Weston, ``Curriculum learning,''
  in \emph{Proceedings of the 26th annual international conference on machine
  learning}, 2009, pp. 41--48.

\bibitem{ram2023context}
O.~Ram, Y.~Levine, I.~Dalmedigos, D.~Muhlgay, A.~Shashua, K.~Leyton-Brown, and
  Y.~Shoham, ``In-context retrieval-augmented language models,''
  \emph{Transactions of the Association for Computational Linguistics},
  vol.~11, pp. 1316--1331, 2023.

\bibitem{kim2024sure}
\BIBentryALTinterwordspacing
J.~Kim, J.~Nam, S.~Mo, J.~Park, S.-W. Lee, M.~Seo, J.-W. Ha, and J.~Shin,
  ``Sure: Summarizing retrievals using answer candidates for open-domain {QA}
  of {LLM}s,'' in \emph{The Twelfth International Conference on Learning
  Representations}, 2024. [Online]. Available:
  \url{https://openreview.net/forum?id=w4DW6qkRmt}
\BIBentrySTDinterwordspacing

\bibitem{10.5555/3618408.3618698}
M.~De~Jong, Y.~Zemlyanskiy, N.~FitzGerald, J.~Ainslie, S.~Sanghai, F.~Sha, and
  W.~W. Cohen, ``Pre-computed memory or on-the-fly encoding? a hybrid approach
  to retrieval augmentation makes the most of your compute,'' in
  \emph{Proceedings of the 40th International Conference on Machine Learning},
  ser. ICML'23.\hskip 1em plus 0.5em minus 0.4em\relax JMLR.org, 2023.

\bibitem{dejong2023glimmergeneralizedlateinteractionmemory}
\BIBentryALTinterwordspacing
M.~de~Jong, Y.~Zemlyanskiy, N.~FitzGerald, S.~Sanghai, W.~W. Cohen, and
  J.~Ainslie, ``Glimmer: generalized late-interaction memory reranker,'' 2023.
  [Online]. Available: \url{https://arxiv.org/abs/2306.10231}
\BIBentrySTDinterwordspacing

\bibitem{asai2024selfrag}
\BIBentryALTinterwordspacing
A.~Asai, Z.~Wu, Y.~Wang, A.~Sil, and H.~Hajishirzi, ``Self-{RAG}: Learning to
  retrieve, generate, and critique through self-reflection,'' in \emph{The
  Twelfth International Conference on Learning Representations}, 2024.
  [Online]. Available: \url{https://openreview.net/forum?id=hSyW5go0v8}
\BIBentrySTDinterwordspacing

\bibitem{10.1007/s11263-022-01611-x}
\BIBentryALTinterwordspacing
P.~Soviany, R.~T. Ionescu, P.~Rota, and N.~Sebe, ``Curriculum learning: A
  survey,'' \emph{Int. J. Comput. Vision}, vol. 130, no.~6, p. 1526–1565,
  Jun. 2022. [Online]. Available:
  \url{https://doi.org/10.1007/s11263-022-01611-x}
\BIBentrySTDinterwordspacing

\bibitem{guo2018curriculumnet}
S.~Guo, W.~Huang, H.~Zhang, C.~Zhuang, D.~Dong, M.~R. Scott, and D.~Huang,
  ``Curriculumnet: Weakly supervised learning from large-scale web images,'' in
  \emph{Proceedings of the European conference on computer vision (ECCV)},
  2018, pp. 135--150.

\bibitem{hacohen2019power}
G.~Hacohen and D.~Weinshall, ``On the power of curriculum learning in training
  deep networks,'' in \emph{International conference on machine
  learning}.\hskip 1em plus 0.5em minus 0.4em\relax PMLR, 2019, pp. 2535--2544.

\bibitem{chen2023multi}
G.~Chen, R.~Zhan, D.~F. Wong, and L.~S. Chao, ``Multi-level curriculum learning
  for multi-turn dialogue generation,'' \emph{IEEE/ACM Transactions on Audio,
  Speech, and Language Processing}, vol.~31, pp. 3958--3967, 2023.

\bibitem{ranjan2017curriculum}
S.~Ranjan and J.~H. Hansen, ``Curriculum learning based approaches for noise
  robust speaker recognition,'' \emph{IEEE/ACM Transactions on Audio, Speech,
  and Language Processing}, vol.~26, no.~1, pp. 197--210, 2017.

\bibitem{platanios-etal-2019-competence}
\BIBentryALTinterwordspacing
E.~A. Platanios, O.~Stretcu, G.~Neubig, B.~Poczos, and T.~Mitchell,
  ``Competence-based curriculum learning for neural machine translation,'' in
  \emph{Proceedings of the 2019 Conference of the North {A}merican Chapter of
  the Association for Computational Linguistics: Human Language Technologies,
  Volume 1 (Long and Short Papers)}, J.~Burstein, C.~Doran, and T.~Solorio,
  Eds.\hskip 1em plus 0.5em minus 0.4em\relax Minneapolis, Minnesota:
  Association for Computational Linguistics, Jun. 2019, pp. 1162--1172.
  [Online]. Available: \url{https://aclanthology.org/N19-1119/}
\BIBentrySTDinterwordspacing

\bibitem{surkov-etal-2022-data}
\BIBentryALTinterwordspacing
M.~Surkov, V.~Mosin, and I.~P. Yamshchikov, ``Do data-based curricula work?''
  in \emph{Proceedings of the Third Workshop on Insights from Negative Results
  in NLP}, S.~Tafreshi, J.~Sedoc, A.~Rogers, A.~Drozd, A.~Rumshisky, and
  A.~Akula, Eds.\hskip 1em plus 0.5em minus 0.4em\relax Dublin, Ireland:
  Association for Computational Linguistics, May 2022, pp. 119--128. [Online].
  Available: \url{https://aclanthology.org/2022.insights-1.16/}
\BIBentrySTDinterwordspacing

\bibitem{xu2020curriculum}
B.~Xu, L.~Zhang, Z.~Mao, Q.~Wang, H.~Xie, and Y.~Zhang, ``Curriculum learning
  for natural language understanding,'' in \emph{Proceedings of the 58th Annual
  Meeting of the Association for Computational Linguistics}, 2020, pp.
  6095--6104.

\bibitem{feng-etal-2025-pretrained}
\BIBentryALTinterwordspacing
Q.~Feng, Y.~Liu, and H.~Schuetze, ``Your pretrained model tells the difficulty
  itself: A self-adaptive curriculum learning paradigm for natural language
  understanding,'' in \emph{Proceedings of the 63rd Annual Meeting of the
  Association for Computational Linguistics (Volume 4: Student Research
  Workshop)}, J.~Zhao, M.~Wang, and Z.~Liu, Eds.\hskip 1em plus 0.5em minus
  0.4em\relax Vienna, Austria: Association for Computational Linguistics, Jul.
  2025, pp. 222--239. [Online]. Available:
  \url{https://aclanthology.org/2025.acl-srw.15/}
\BIBentrySTDinterwordspacing

\bibitem{nair-etal-2024-curriculum}
\BIBentryALTinterwordspacing
M.~Na{\"i}r, K.~Yamani, L.~Lhadj, and R.~Baghdadi, ``Curriculum learning for
  small code language models,'' in \emph{Proceedings of the 62nd Annual Meeting
  of the Association for Computational Linguistics (Volume 4: Student Research
  Workshop)}, X.~Fu and E.~Fleisig, Eds.\hskip 1em plus 0.5em minus 0.4em\relax
  Bangkok, Thailand: Association for Computational Linguistics, Aug. 2024, pp.
  390--401. [Online]. Available:
  \url{https://aclanthology.org/2024.acl-srw.44/}
\BIBentrySTDinterwordspacing

\bibitem{wang-etal-2025-knowledge}
\BIBentryALTinterwordspacing
H.~Wang, M.~Nuo, and S.~Jiang, ``Knowledge graph entity typing with curriculum
  contrastive learning,'' in \emph{Proceedings of the 31st International
  Conference on Computational Linguistics}, O.~Rambow, L.~Wanner,
  M.~Apidianaki, H.~Al-Khalifa, B.~D. Eugenio, and S.~Schockaert, Eds.\hskip
  1em plus 0.5em minus 0.4em\relax Abu Dhabi, UAE: Association for
  Computational Linguistics, Jan. 2025, pp. 574--583. [Online]. Available:
  \url{https://aclanthology.org/2025.coling-main.38/}
\BIBentrySTDinterwordspacing

\bibitem{Zhu2022FromET}
\BIBentryALTinterwordspacing
Y.~Zhu, J.~Nie, Y.~Su, H.~Chen, X.~Zhang, and Z.~Dou, ``From easy to hard: A
  dual curriculum learning framework for context-aware document ranking,''
  \emph{Proceedings of the 31st ACM International Conference on Information \&
  Knowledge Management}, 2022. [Online]. Available:
  \url{https://api.semanticscholar.org/CorpusID:251719021}
\BIBentrySTDinterwordspacing

\bibitem{zeng2022curriculum}
H.~Zeng, H.~Zamani, and V.~Vinay, ``Curriculum learning for dense retrieval
  distillation,'' in \emph{Proceedings of the 45th International ACM SIGIR
  Conference on Research and Development in Information Retrieval}, 2022, pp.
  1979--1983.

\bibitem{he2023capstone}
X.~He, Y.~Gong, A.-L. Jin, H.~Zhang, A.~Dong, J.~Jiao, S.~Yiu, and N.~Duan,
  ``Capstone: Curriculum sampling for dense retrieval with document
  expansion,'' in \emph{Proceedings of the 2023 Conference on Empirical Methods
  in Natural Language Processing}, 2023, pp. 10\,531--10\,541.

\bibitem{fisch2019mrqa}
A.~Fisch, A.~Talmor, R.~Jia, M.~Seo, E.~Choi, and D.~Chen, ``{MRQA} 2019 shared
  task: Evaluating generalization in reading comprehension,'' in
  \emph{Proceedings of 2nd Machine Reading for Reading Comprehension (MRQA)
  Workshop at EMNLP}, 2019.

\bibitem{10.1162/tacl_a_00276}
\BIBentryALTinterwordspacing
T.~Kwiatkowski, J.~Palomaki, O.~Redfield, M.~Collins, A.~Parikh, C.~Alberti,
  D.~Epstein, I.~Polosukhin, J.~Devlin, K.~Lee, K.~Toutanova, L.~Jones,
  M.~Kelcey, M.-W. Chang, A.~M. Dai, J.~Uszkoreit, Q.~Le, and S.~Petrov,
  ``Natural questions: A benchmark for question answering research,''
  \emph{Transactions of the Association for Computational Linguistics}, vol.~7,
  pp. 453--466, 08 2019. [Online]. Available:
  \url{https://doi.org/10.1162/tacl\_a\_00276}
\BIBentrySTDinterwordspacing

\bibitem{joshi-etal-2017-triviaqa}
\BIBentryALTinterwordspacing
M.~Joshi, E.~Choi, D.~Weld, and L.~Zettlemoyer, ``{T}rivia{QA}: A large scale
  distantly supervised challenge dataset for reading comprehension,'' in
  \emph{Proceedings of the 55th Annual Meeting of the Association for
  Computational Linguistics (Volume 1: Long Papers)}, R.~Barzilay and M.-Y.
  Kan, Eds.\hskip 1em plus 0.5em minus 0.4em\relax Vancouver, Canada:
  Association for Computational Linguistics, Jul. 2017, pp. 1601--1611.
  [Online]. Available: \url{https://aclanthology.org/P17-1147/}
\BIBentrySTDinterwordspacing

\bibitem{mallen-etal-2023-trust}
\BIBentryALTinterwordspacing
A.~Mallen, A.~Asai, V.~Zhong, R.~Das, D.~Khashabi, and H.~Hajishirzi, ``When
  not to trust language models: Investigating effectiveness of parametric and
  non-parametric memories,'' in \emph{Proceedings of the 61st Annual Meeting of
  the Association for Computational Linguistics (Volume 1: Long Papers)},
  A.~Rogers, J.~Boyd-Graber, and N.~Okazaki, Eds.\hskip 1em plus 0.5em minus
  0.4em\relax Toronto, Canada: Association for Computational Linguistics, Jul.
  2023, pp. 9802--9822. [Online]. Available:
  \url{https://aclanthology.org/2023.acl-long.546/}
\BIBentrySTDinterwordspacing

\bibitem{yang-etal-2018-hotpotqa}
\BIBentryALTinterwordspacing
Z.~Yang, P.~Qi, S.~Zhang, Y.~Bengio, W.~Cohen, R.~Salakhutdinov, and C.~D.
  Manning, ``{H}otpot{QA}: A dataset for diverse, explainable multi-hop
  question answering,'' in \emph{Proceedings of the 2018 Conference on
  Empirical Methods in Natural Language Processing}, E.~Riloff, D.~Chiang,
  J.~Hockenmaier, and J.~Tsujii, Eds.\hskip 1em plus 0.5em minus 0.4em\relax
  Brussels, Belgium: Association for Computational Linguistics, Oct.-Nov. 2018,
  pp. 2369--2380. [Online]. Available: \url{https://aclanthology.org/D18-1259/}
\BIBentrySTDinterwordspacing

\bibitem{ho-etal-2020-constructing}
\BIBentryALTinterwordspacing
X.~Ho, A.-K. Duong~Nguyen, S.~Sugawara, and A.~Aizawa, ``Constructing a
  multi-hop {QA} dataset for comprehensive evaluation of reasoning steps,'' in
  \emph{Proceedings of the 28th International Conference on Computational
  Linguistics}, D.~Scott, N.~Bel, and C.~Zong, Eds.\hskip 1em plus 0.5em minus
  0.4em\relax Barcelona, Spain (Online): International Committee on
  Computational Linguistics, Dec. 2020, pp. 6609--6625. [Online]. Available:
  \url{https://aclanthology.org/2020.coling-main.580/}
\BIBentrySTDinterwordspacing

\bibitem{izacardunsupervised}
G.~Izacard, M.~Caron, L.~Hosseini, S.~Riedel, P.~Bojanowski, A.~Joulin, and
  E.~Grave, ``Unsupervised dense information retrieval with contrastive
  learning,'' \emph{Transactions on Machine Learning Research}, 2021.

\bibitem{petroni-etal-2021-kilt}
\BIBentryALTinterwordspacing
F.~Petroni, A.~Piktus, A.~Fan, P.~Lewis, M.~Yazdani, N.~De~Cao, J.~Thorne,
  Y.~Jernite, V.~Karpukhin, J.~Maillard, V.~Plachouras, T.~Rockt{\"a}schel, and
  S.~Riedel, ``{KILT}: a benchmark for knowledge intensive language tasks,'' in
  \emph{Proceedings of the 2021 Conference of the North American Chapter of the
  Association for Computational Linguistics: Human Language
  Technologies}.\hskip 1em plus 0.5em minus 0.4em\relax Online: Association for
  Computational Linguistics, Jun. 2021, pp. 2523--2544. [Online]. Available:
  \url{https://aclanthology.org/2021.naacl-main.200}
\BIBentrySTDinterwordspacing

\bibitem{qwen2.5}
\BIBentryALTinterwordspacing
Q.~Team, ``Qwen2.5: A party of foundation models,'' September 2024. [Online].
  Available: \url{https://qwenlm.github.io/blog/qwen2.5/}
\BIBentrySTDinterwordspacing

\bibitem{zheng2024llamafactory}
\BIBentryALTinterwordspacing
Y.~Zheng, R.~Zhang, J.~Zhang, Y.~Ye, Z.~Luo, Z.~Feng, and Y.~Ma,
  ``Llamafactory: Unified efficient fine-tuning of 100+ language models,'' in
  \emph{Proceedings of the 62nd Annual Meeting of the Association for
  Computational Linguistics (Volume 3: System Demonstrations)}.\hskip 1em plus
  0.5em minus 0.4em\relax Bangkok, Thailand: Association for Computational
  Linguistics, 2024. [Online]. Available: \url{http://arxiv.org/abs/2403.13372}
\BIBentrySTDinterwordspacing

\end{thebibliography}

\vspace{-33pt}
\begin{IEEEbiography}[{\includegraphics[width=1in,height=1.25in,clip,keepaspectratio]{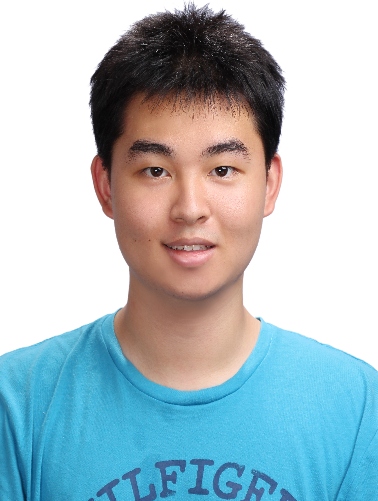}}]{Shaohan Wang} received the B.Sc. degree in electronic information engineering from the University of Science and Technology of China (USTC), Hefei, China, in 2023. He is currently a Ph.D. student in the School of Cyberspace Science and Technology, USTC. His research interests include natural language processing and large language model. 
\end{IEEEbiography}

\vspace{-33pt}

\begin{IEEEbiography}[{\includegraphics[width=1in,height=1.25in,clip,keepaspectratio]{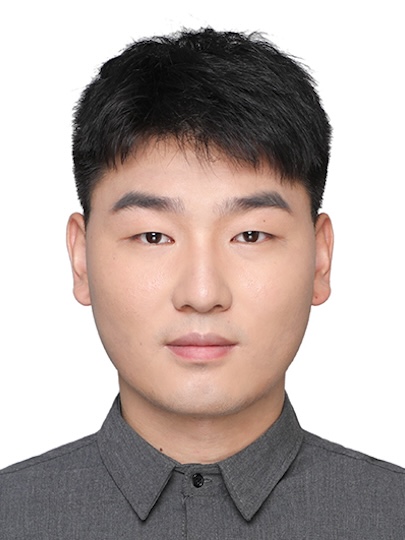}}]{Licheng Zhang} received the B.Sc. degree in electronic information engineering from the University of Science and Technology of China (USTC), Hefei, China, in 2018. He is currently a last year Ph.D. student in the Department of Electronic Engineering and Information Science, USTC. His research interests include natural language processing and large language model.
\end{IEEEbiography}

\vspace{-33pt}

\begin{IEEEbiography}[{\includegraphics[width=1in,height=1.25in,clip,keepaspectratio]{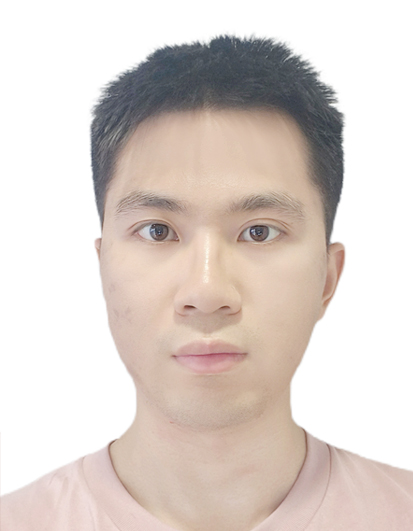}}]{Zheren Fu} received the Ph.D. and B.Sc. degrees from the University of Science and Technology of China, Hefei, China, in 2025 and 2020, respectively. He is currently a postdoctoral researcher with the School of Information Science and Technology, University of Science and Technology of China, Hefei, China. His research interests mainly cover LLM, vision-language models, and cross-modal alignment.
\end{IEEEbiography}

\vspace{-33pt}

\begin{IEEEbiography}[{\includegraphics[width=1in,height=1.25in,clip,keepaspectratio]{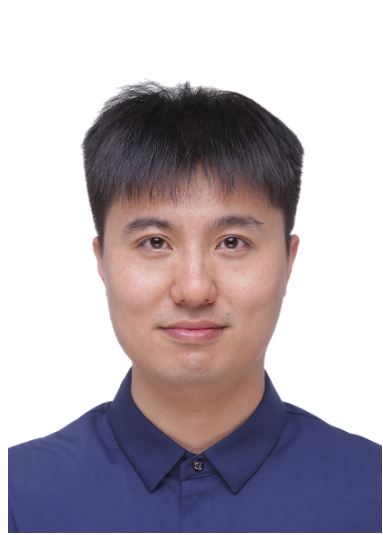}}]{Zhendong Mao} (M’23) received the Ph.D. degree in computer application technology from the Institute of Computing Technology, Chinese Academy of Sciences, in 2014. He is currently a professor with the School of Cyberspace Science and Technology, University of Science and Technology of China, Hefei, China. He was an assistant professor with the Institute of Information Engineering, Chinese Academy of Sciences, Beijing, from 2014 to 2018. His research interests include the fields of LLM, cross-modal understanding and generation. He serves as an Associate Editor of the IEEE TRANSACTIONS ON CIRCUITS AND SYSTEMS
FOR VIDEO TECHNOLOGY and IEEE TRANSACTIONS ON MULTIMEDIA.
\end{IEEEbiography}

\vspace{-33pt}

\begin{IEEEbiography}[{\includegraphics[width=1in,height=1.25in,clip,keepaspectratio]{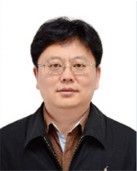}}]{Yongdong Zhang} (M’08–SM’13-F’24) received the Ph.D. degree in electronic engineering from Tianjin University, Tianjin, China, in 2002. He is currently a Professor with the School of Information Science and Technology, University of Science and Technology of China. His current research interests are in the fields of multimedia content analysis and understanding, multimedia content security, video encoding, and streaming media technology. He has authored over 200 refereed journal and conference papers, accumulating more than 33,000 citations on Google Scholar. He was a recipient of the best paper awards in PCM 2013, ICIMCS 2013, ICME 2010, the best student paper award in ACM Multimedia 2022 and the Best Paper Candidate in ICME 2011. He serves as an Editorial Board Member of the Multimedia Systems Journal and the IEEE TRANSACTIONS ON MULTIMEDIA. He is a fellow of the IEEE.
\end{IEEEbiography}




\end{document}